\documentclass{article} 
\usepackage{iclr2027_conference,times}

\usepackage{amsmath,amsfonts,bm}

\def\eqref#1{equation~\ref{#1}}

\def\1{\bm{1}}

\DeclareMathAlphabet{\mathsfit}{\encodingdefault}{\sfdefault}{m}{sl}
\SetMathAlphabet{\mathsfit}{bold}{\encodingdefault}{\sfdefault}{bx}{n}

\usepackage{hyperref}
\usepackage{url}
\usepackage{graphicx}
\usepackage[most]{tcolorbox}
\usepackage{xcolor}
\usepackage{fancyvrb}
\usepackage{natbib}
\usepackage{amsmath}
\usepackage{amssymb}
\usepackage{ragged2e}
\usepackage{enumitem}
\usepackage{array}
\usepackage{booktabs} 
\usepackage{float}
\usepackage{wrapfig}
\usepackage{makecell}
\usepackage{placeins}
\usepackage{subcaption}
\usepackage[T1]{fontenc}
\usepackage{threeparttable}

\DeclareRobustCommand{\curlyamp}{%
  {\fontencoding{T1}%
   \fontfamily{EBGaramond-OsF}%
   \fontseries{m}%
   \fontshape{it}%
   \selectfont
   \&}%
}
\title{On Reasoning Correctness Directions}

\author{\textbf{Munachiso Samuel Nwadike}$^{1,3}$ \quad
\textbf{Zayd M. K. Zuhri}$^{2}$ \\
\textbf{Rifo Genadi}$^{1}$ \quad
\textbf{Abzal Nurgazy}$^{1}$ \quad
\textbf{Ali Mekky}$^{1}$ \quad
\textbf{Alham Fikri Aji}$^{1}$ \\
\textbf{Zangir Iklassov}$^{1}$ \quad
\textbf{Kentaro Inui}$^{1,3,4}$ \\[0.5em]
$^{1}$MBZUAI, UAE \quad
$^{2}$Mila, Canada \quad
$^{3}$RIKEN AIP, Japan \quad
$^{4}$Tohoku University, Japan
\\
\texttt{munachiso.nwadike@mbzuai.ac.ae}
}

\newcommand{\tokenlyfont}[1]{%
  \begingroup
  \renewcommand{\ttdefault}{lmtt}
  \normalfont\ttfamily\mdseries
  \colorbox{gray!12}{#1}%
  \endgroup
}

\newcommand{\tokenlynogray}[1]{%
  \begingroup
  \renewcommand{\ttdefault}{lmtt}
  \ttfamily\mdseries
  #1%
  \endgroup
}

\newcommand{\tablethreefont}{%
  \fontsize{8.5pt}{9.5pt}\selectfont
}

\definecolor{exampletitle}{RGB}{28, 69, 135}   
\definecolor{examplebg}{RGB}{242, 246, 255}    

\newcounter{heuristiccounter}
\newtcolorbox{heuristicslist}[1][]{%
  enhanced,
  breakable,
  colback=examplebg,
  frame hidden,
  arc=0mm,
  sharp corners,
  left=10pt,
  right=10pt,
  top=10pt,
  bottom=10pt,
  before skip=10pt,
  after skip=10pt,
  parbox=false,
  fontupper=\small\justifying,
  before upper={\setcounter{heuristiccounter}{0}}, 
  #1
}
\newcounter{heuristicsubcounter}[heuristiccounter]  
\renewcommand{\theheuristicsubcounter}{\theheuristiccounter\alph{heuristicsubcounter}} 
\newcommand{\heuristicitem}[1]{%
  \par\addvspace{1.5em}
  \refstepcounter{heuristiccounter}
  \setcounter{heuristicsubcounter}{0} %
  \noindent\textcolor{exampletitle}{\large{\theheuristiccounter. #1}}\\[0.5em]
  \ignorespaces
}
\newcommand{\heuristicsubitem}[1]{%
  \par\addvspace{1em}
  \refstepcounter{heuristicsubcounter} %
  \noindent\textcolor{exampletitle}{{\theheuristicsubcounter. #1}} %
  \ignorespaces
}

\newcommand{\raisedxmark}{\raisebox{-0.45ex}{\includegraphics[height=1em]{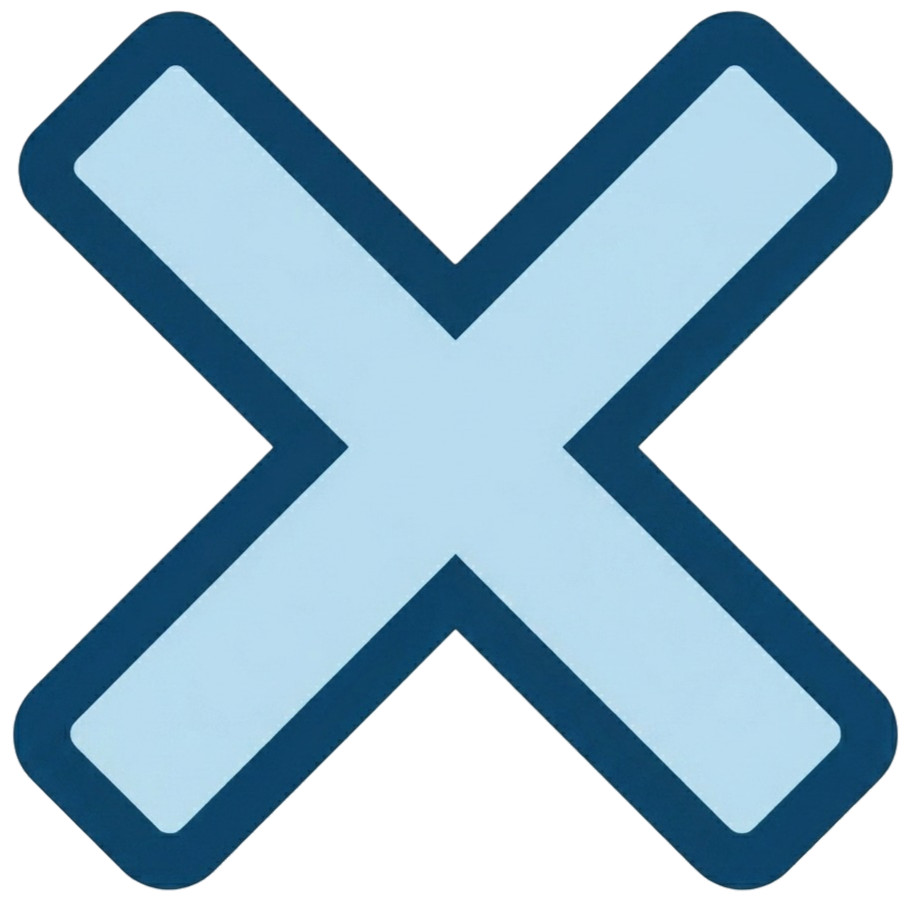}}}
\newcommand{\raisedcheckmark}{\raisebox{-0.45ex}{\includegraphics[height=1em]{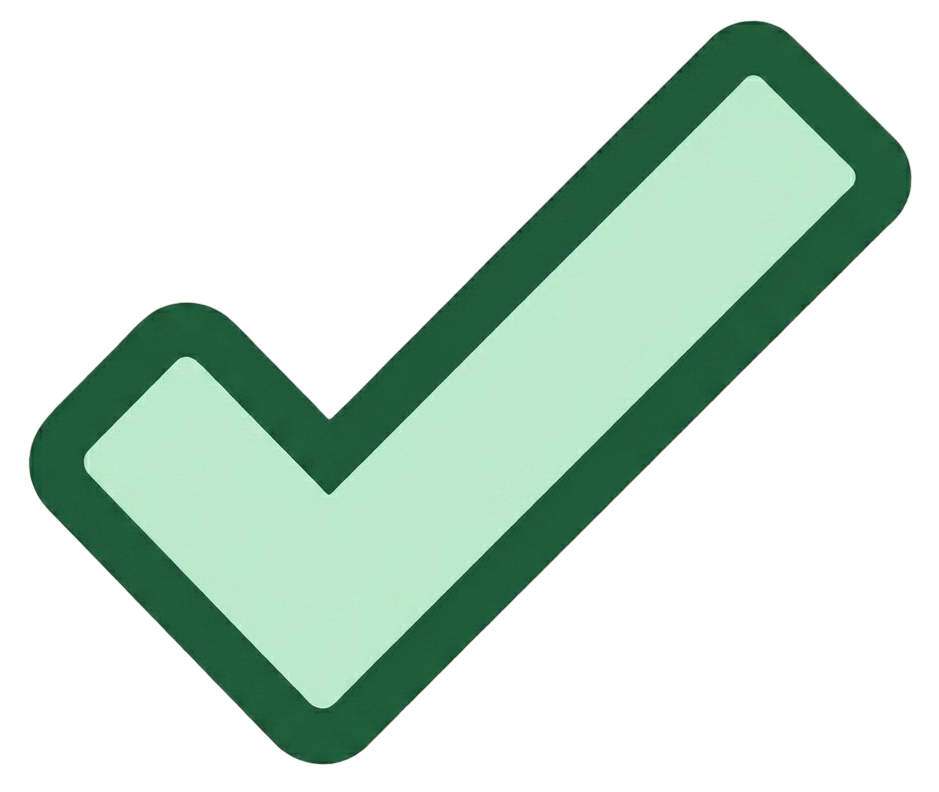}}}

\definecolor{chainexamplebodybg}{HTML}{F2F6FF}
\definecolor{chainexampleheaderbg}{HTML}{F2F6FF} 
\definecolor{chainexampleframe}{HTML}{1C4587}
\definecolor{chainexampleheadertext}{HTML}{1C4587}

\newtcolorbox[use counter=chainexample]{chainexample}[2][]{%
  enhanced,
  colback=chainexamplebodybg,
  colbacktitle=chainexampleheaderbg,
  coltitle=chainexampleheadertext,
  colframe=chainexampleframe!45,
  boxrule=0.45pt,
  arc=0.0mm,
  left=6pt,
  right=6pt,
  top=5pt,
  bottom=5pt,
  before skip=4pt,
  after skip=4pt,
  fonttitle=\small,
  title={Example~\thetcbcounter.\ #2},
#1
}

\newtcolorbox[use counter=chainexample]{chainexamplebreakable}[2][]{%
  enhanced,
  breakable,
  colback=chainexamplebodybg,
  colbacktitle=chainexampleheaderbg,
  coltitle=chainexampleheadertext,
  colframe=chainexampleframe!45,
  boxrule=0.45pt,
  arc=0.0mm,
  left=6pt,
  right=6pt,
  top=5pt,
  bottom=5pt,
  before skip=4pt,
  after skip=4pt,
  fonttitle=\small,
  title={Example~\thetcbcounter.\ #2},
#1
}

\newcommand{\chaintracefontsize}{%
  \fontsize{7.8pt}{9pt}\selectfont
}

\definecolor{pretrainpastel}{HTML}{FFFFFF}
\definecolor{finetunepastel}{HTML}{FFFFFF}

\iclrfinalcopy 
\begin{document}

\maketitle

\begin{abstract}
We present experimental evidence suggesting that information about whether a model’s reasoning is on the right track can be linearly decodable without defining a causally useful direction for steering. To study this, we introduce \tokenlynogray{ChainSearch}, a controlled arithmetic reasoning task that extends Countdown with explicit search and backtracking. We train a 3-\textbf{\underline{m}}illion-parameter model to solve this task. Its compact size allows us to study internal model correctness directions in detail using linear probes, mean activation differences, and direct optimisation, as part of a rigorous search for internal reasoning-correctness directions. We find that while the model contains statistically significant internal directions of reasoning correctness, none of the resulting directions consistently improves reasoning under causal intervention. This calls into question whether the linear representation hypothesis is sufficient for identifying causally useful directions in the context of reasoning. Our code is available at \url{https://github.com/samunaai/chainsearch}.
\end{abstract}

\section{Introduction}
\label{sec:intro}

\begin{wrapfigure}[21]{r}{0.6\textwidth}
    \centering
    \vspace{-0.7em}
    \includegraphics[width=\linewidth]{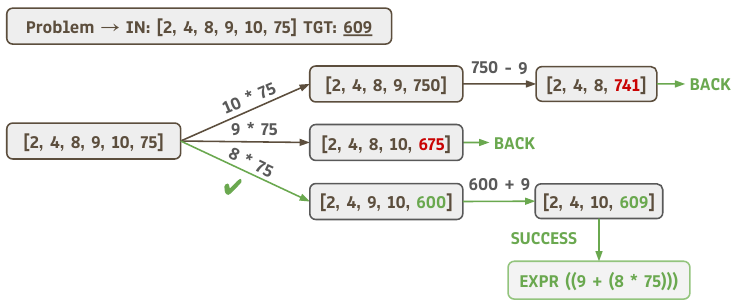}
    \vspace{-1.6em}
    \caption{Understanding how reasoning models arrive at their answers has attracted growing interest (Appendix ~\ref{sec:related_work}). Reasoning often entails exploring alternative paths, identifying dead ends, and backtracking to earlier states. \protect\tokenlynogray{ChainSearch} makes this process explicit in an arithmetic game, where each step can preserve or remove the possibility of reaching the target. Red numerical results mark states from which the target can no longer be reached without backtracking, while green results mark states where it remains reachable. \texttt{BACK} restores an earlier state for further exploration, and \texttt{SUCCESS} marks attainment of the target.}
    \label{fig:graph_diagram}
\end{wrapfigure}
Empirical research on human metacognition suggests that, in structured mathematical problem solving, people can monitor their progress toward a solution, with such judgments predicting subsequent performance \citep{metcalfe1987intuition}. The same holds true in human decision making \citep{yeung2012metacognition}. This raises an interesting question of whether reasoning in language models, much like other properties \citep{li2023inference,arditi2024refusal,genadi-etal-2026-sycophancy}, bears hallmarks of the \textit{linear representation hypothesis} \citep{park2024linear}.

Recent work also finds that linear probes can predict
whether intermediate answers are correct from internal
representations \citep{zhang2025reasoning}, but does not investigate causal steering specifically for reasoning correctness. 
Although prior work has documented limitations in the reliability of linear steering \citep{tan2024,braun2025}, interventions in reasoning models have successfully modulated behaviours such as uncertainty and backtracking \citep{venhoff2025understanding}, while steering directions encoding behaviours such as {reflection and transition} have improved answer accuracy \citep{chen2025seal,zhuang2026reliable}. These studies show that controlling particular reasoning behaviours can \textit{\underline{indirectly}} improve answer accuracy. However, \textit{they do not establish whether correctness itself} is encoded in a \textit{\underline{causally} useful direction} that can reliably guide a model towards the correct solution.

One reason why answering this particular question is challenging, is that reasoning correctness must be examined throughout an evolving trace. Compared to properties such as truthfulness and sycophancy, which can be evaluated after an entire complete response has been produced \citep{li2023inference,genadi-etal-2026-sycophancy}, reasoning correctness can \textit{change at multiple token positions within a single trace} (e.g. due to backtracking). In other words, we must track whether intermediate states still permit a correct solution through forward operations, or whether the model needs to backtrack to an earlier state from which a solution remains possible. This complicates the choice of where to check for correctness directions, or to steer the model, as there are a wealth of candidate token positions.

This question of when to intervene is accompanied by a localisation problem within the model itself. For reasoning, we must determine not only at which token position a correctness signal is examined along the reasoning trace, but also \textit{where it appears across layers and components}. \citet{zhang2025reasoning} use final-layer correctness signals to decide when to stop generation. However, by the time correctness is detected at this layer, the preceding attention and MLP computations have already taken place, making the final layer a limited intervention site for modifying the internal computations that produced the signal. It is therefore more desirable, when studying reasoning, to be able to investigate directions recovered from deeper parts of the model such as attention heads, residual streams, and MLPs. The resulting number of probe evaluations grows multiplicatively with dataset size, trace length, model depth, and the number of attention heads examined. \textit{Even at the scale of a few billion parameters, dense probing and causal steering can become computationally prohibitive}, particularly when each intervention must be evaluated by generating a new response.  

To make such investigation tractable, we draw inspiration from \citet{lee2024teaching}, \citet{ye2024physics}, and \citet{nwadike2025recall}, who constrain the model and training data to train compact models specialised for particular tasks to isolate particular model capabilities. We construct a similarly controlled setting for reasoning, which we name \tokenlynogray{ChainSearch}. In this setting, we train compact models \textit{from scratch}, giving us full control over their training, while allowing us to probe correctness signals and test interventions across token positions, layers, and components.

Our main contributions are the following: (1) We introduce \tokenlynogray{ChainSearch} (see Section~\ref{sec:task_formulation}), a controllable search-based CoT sandbox for studying reasoning correctness in a computationally efficient way. It uses compact transformers trained from scratch on Countdown traces generated by a modified form of the A* algorithm (see Appendix \ref{sec:why_astar}) with explicit state transitions and backtracking.  
(2) We propose a probing setup that studies reasoning
signals across token positions, layers, and model components.
This lets us test whether models encode local information
about correctness, search direction, state progression,
and backtracking. (3) We use \tokenlynogray{ChainSearch} to test whether reasoning models know when they are on the right track across both successful and failed search trajectories. We find that although correctness is linearly decodable, the recovered directions do not consistently improve reasoning under causal intervention, even when the vectors are directly optimised to increase the likelihood of correct final answers.


 

\vspace{-0.1em}
\section{ChainSearch Task}
\label{sec:task_formulation}
\vspace{-0.1em}

\begin{SaveVerbatim}[commandchars=\\\{\}]{chaintrace_example_one}
\underline{Problem}
IN: [ 3, 6, 10, 10, 50, 75 ] TGT: 425

\underline{Solution}
[ 3, 6, 10, 10, 50, 75 ] -> 6 * 75 = 450
-> [ 3, 10, 10, 50, 450 ] ; \raisedxmark
BACK -> [ 3, 6, 10, 10, 50, 75 ] ;
[ 3, 6, 10, 10, 50, 75 ] -> 10 * 50 = 500
-> [ 3, 6, 10, 75, 500 ] ;
[ 3, 6, 10, 75, 500 ] -> 500 - 75 = 425
-> [ 3, 6, 10, 425 ] ;
SUCCESS ;
EXPR (((10 * 50) - 75)) \raisedcheckmark
\end{SaveVerbatim}

\vspace{-0.2em}


\begin{wrapfigure}[14]{r}{0.53\textwidth}
\vspace{-1.3em}
\begin{chainexample}[label={example:first_sample_trace}]{A \tokenlynogray{ChainSearch} training trace}
\UseVerbatim[fontsize=\chaintracefontsize]{chaintrace_example_one}
\end{chainexample}
\end{wrapfigure} 
\leavevmode\tokenlynogray{ChainSearch} is a controlled arithmetic search testbed based on the Countdown numbers puzzle~\citep{alliot2015final,colton2014countdown}. The player is given a \textit{\textbf{target}} number between \(\texttt{100}\) and \(\texttt{999}\), as well as \textit{\textbf{six inputs}} drawn from a pool of 24 numbers (\(\texttt{1}\) to \(\texttt{10}\) repeated twice, together with \(\texttt{25}\), \(\texttt{50}\), \(\texttt{75}\), and \(\texttt{100}\)). The goal is to combine the inputs to reach the target. 

A legal move is one that selects two values, applies one \textit{operator} from \(\{+, -, \times, \div\}\), and uses the result, combined with other numbers remaining, to perform a \textit{state transition}, replacing the two numbers with the result as progress towards the target. For instance, in Example~\ref{example:first_sample_trace}, after multiplying \(6\) and \(75\) to obtain \(450\), we transition from \(\texttt{[3,6,10,10,50,75]}\) to \(\texttt{[3,10,10,50,450]}\).  Here, an \textit{arithmetic operation} is the calculation applied to two values (as opposed to the operator itself), and a \textit{state transition} is the complete update from the preceding state to the resulting state. A ``\texttt{BACK} transition'' restores an earlier state.

In \tokenlynogray{ChainSearch}, there is no requirement to use all the inputs to attain the target, but each input may be used only once, and only positive integers can be produced. The final answer is not the target itself, but the ``expression'' (\(\texttt{EXPR}\) in Example ~\ref{example:first_sample_trace}) at the end, which should evaluate to the target.

The difference between Countdown and \tokenlynogray{ChainSearch} is that \tokenlynogray{ChainSearch} introduces an explicit representation of backtracking. Countdown, despite being extensively studied in mathematical and computational settings (see Appendix~\ref{chainsearch_complexity}) was originally developed as a game for humans. To test reasoning in a language model, we introduce \texttt{BACK} tokens to mark when the search abandons a branch and restores an earlier state. 

We generate \tokenlynogray{ChainSearch} search trajectories using a modified form of A\textsuperscript{*} guided by human-inspired Countdown heuristics (see Appendix~\ref{sec:heuristics_list}, which details the heuristics used, and a gentler introduction in Appendix~\ref{sec:why_astar}). Our autoregressive decoder-only transformer is trained on such data (as shown in Example~\ref{example:first_sample_trace}), to produce the entire trace, including the final expression, conditioned only on the ``input numbers \(\texttt{IN}\) \texorpdfstring{\curlyamp}{\&} target \(\texttt{TGT}\)''. Further examples containing different numbers of \texttt{BACK} transitions are provided in Appendix~\ref{sec:additional_trace_examples}. At a high level, the SFT split of \tokenlynogray{ChainSearch} contains 1,016,676 examples.  We additionally generate 6,948,231 exploratory traces for pretraining.  The corresponding data splits are addressed further in Appendix~\ref{sec:dataset_splits}. Additional details on model training and task performance are provided in Appendices~\ref{dataset_details} and~\ref{sec:model_training}.

\section{Correctness Direction Analysis}
\label{sec:experiments}

To investigate whether the model contains causally useful correctness directions, we first specify how candidate directions are defined. We then ask whether correctness is linearly decodable and whether the recovered directions remain stable across various dimensions. We construct and analyse these directions using 115,367 generated traces, including a 16,114-trace test split used in full for held-out probe reporting. Section~\ref{sec:steering_results} provides a subsequent causal test, evaluating whether interventions along these directions improve reasoning. 

\subsection{Defining and Studying Correctness Directions}
\label{sec:correctness_direction_axes}

Each candidate correctness direction is defined by four experimental axes. The \textit{target label} \((\ell)\) specifies the type of correctness, the \textit{semantic/token position} \((p)\) specifies when activations are measured, the \textit{direction method} \((m)\) specifies how the direction is constructed, and the \textit{model component} \((c)\) specifies where inside the model it is measured.  We denote the resulting direction by \(u_{m,\ell,p,c}\) and describe each axis below

\begin{enumerate}[
    label=\Roman*.,
    leftmargin=1.5em,
    labelsep=0.4em,
    itemsep=0.3em,
    topsep=0.3em,
    parsep=0pt,
    partopsep=0pt
]

\item {Target labels \((l)\):}
\begin{enumerate}[
    label=\alph*),
    leftmargin=1.5em,
    labelsep=0.4em,
    itemsep=0.35em,
    topsep=0.25em,
    parsep=0pt,
    partopsep=0pt
]
\item \textit{Reachability.} This records whether the state produced by an arithmetic state transition, or restored by a \tokenlyfont{BACK} transition, can still reach the target using only valid forward arithmetic operations. All selected token positions within an arithmetic state transition share the label of its resulting state, while token positions associated with \tokenlyfont{BACK} share the label of the restored state. This yields an exact intermediate label without requiring explicit intermediate answers.
\item \textit{Final-answer correctness.} This records whether the complete trace produces a correct final expression. It is attached to each selected token position using the trace's eventual outcome.
\end{enumerate}
Utilising these two labels helps distinguish local state viability from eventual success.

\item {Semantic token positions  \((p)\):} We consider semantic token positions that represent \textit{key decision points} within the models reasoning trace. Particularly, we look at the first operand position, the second operand position, the operator tokens \tokenlyfont{$+$}\tokenlyfont{$-$}\tokenlyfont{$\times$}\tokenlyfont{$\div$}, the result token for a given operation, and the resulting state of a state transition. We also look at \(\texttt{BACK}\),  \(\texttt{SUCCESS}\), and \(\texttt{EXPR}\) positions.

\item {Direction methods \((m)\):} 
\begin{enumerate}[
    label=\alph*),
    leftmargin=1.5em,
    labelsep=0.4em,
    itemsep=0.35em,
    topsep=0.25em,
    parsep=0pt,
    partopsep=0pt
]
\item \textit{Linear-probe vectors.}
For each target label, semantic token position, and model component, we train a linear classifier and use its weight vector as the candidate direction.
\item \textit{CAA.} We subtract the negative-class mean activation from the positive-class mean activation. 
\item \textit{Direct Optimisation vectors.}
We freeze the model \(M\) and optimise an additive vector \(v\) by minimising the token-level cross-entropy loss \(\min_v \mathcal{L}_{\mathrm{ans}}(M(h+v))\) over a valid final-answer sequence (additional details on this control are provided in Appendix~\ref{sec:direct_optimisation_reachability}).
\end{enumerate}
We sought to utilise a representative collection of direction vector methods (see Appendix~\ref{sec:types_of_linear_steering}). We unit-normalise these vectors to allow for a fair comparison of steering strength later on.
\item {Model locations \((c)\):} We construct directions at every attention head, residual stream, MLP output, and the final layer-normalised hidden state, following prior work (see Section~\ref{sec:intro} and Appendix~\ref{sec:related_work}).

\end{enumerate}

\subsection{Linear Probes}
\label{sec:probing_interp}
Although we also construct directions using CAA and Direct Optimisation in Section~\ref{sec:cross_method_comparison}, we begin with linear probing as a relatively lightweight supervised test of whether local reachability and final-answer correctness are linearly decodable from the model's activations. We train probes at each semantic token position and model component, selecting all hyperparameters on validation data and reporting performance exclusively on the held-out test split.

\begin{wrapfigure}[22]{r}{0.5\textwidth}
    \centering
    \vspace{-1.1em}
    \includegraphics[width=\linewidth]{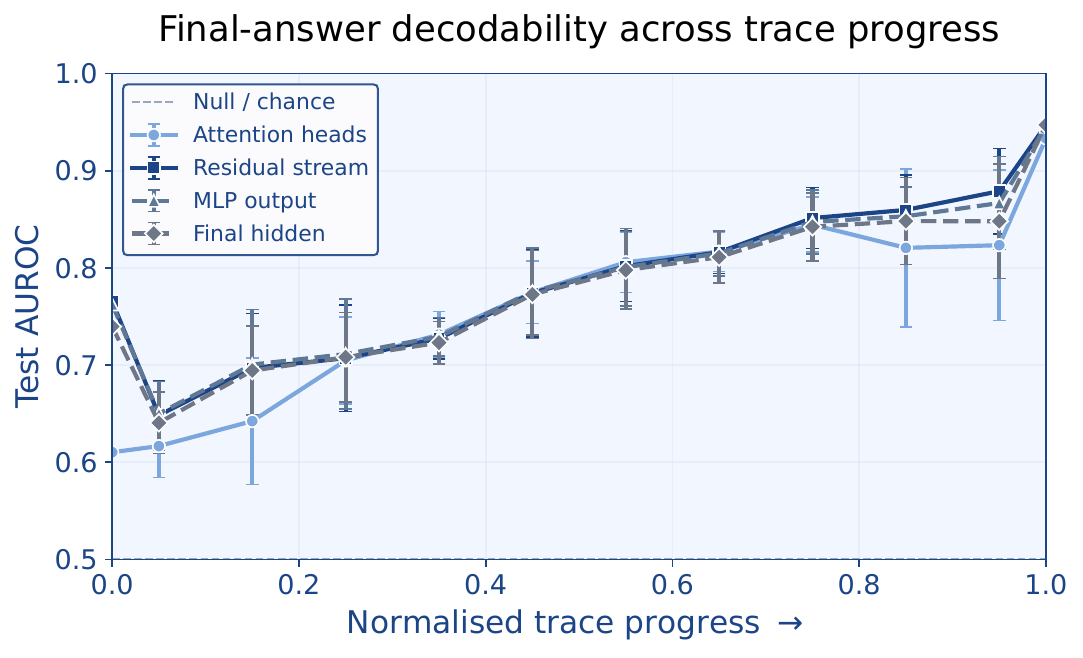}
    \vspace{-1.8em}
    \caption{By normalising reasoning trace lengths, we compare how well linear probes predict final-answer correctness at different stages of generation. Correctness is already decodable well above chance early in the trace and generally becomes more decodable as generation proceeds. This pattern appears across attention heads, residual streams, MLP outputs, and the final hidden representation, with AUROC reaching approximately \(0.94\) by the end. Error bars show one standard deviation around the mean.}
    \label{fig:final_answer_decodability_trace_progress}
\end{wrapfigure} 

Figure~\ref{fig:final_answer_decodability_trace_progress} shows how final-answer decodability develops over the full reasoning trace. Held-out AUROC is already above chance early in generation and, after a small initial dip, steadily increases as the trace progresses. This pattern is similar across all model locations \(c\), although attention heads lag slightly at the starting and ending. 

Because traces containing many \texttt{BACK} transitions contribute more intermediate positions, we repeat the analysis after restricting the data to traces containing at most eight \texttt{BACK} transitions. Appendix ~\ref{sec:appendix_extra_probing_results} Figure~\ref{fig:final_answer_decodability_trace_progress_max_8_backtracks} closely reproduces the trajectory of increasing linear probe AUROC, indicating that the increase towards the end of the trace is not primarily driven by unusually long or heavily backtracking traces.

These results establish that the model's activations contain substantial information about both local reachability and eventual success that can be decoded with a linear probe. However, even if a linear classifier recovers a correctness direction involved in the model's subsequent reasoning, that direction may not best support effective causal steering. We therefore examine whether alternative construction methods recover different candidate directions. Specifically, we construct additional directions using CAA and Direct Optimisation.


\subsection{Beyond Linear Probes (CAA \texorpdfstring{\curlyamp}{\&} Direct Optimisation)}
\label{sec:cross_method_comparison}

Having so far examined correctness through linear probes, we now broaden the search for candidate correctness directions using CAA and Direct Optimisation. This allows us to have several methods for subsequent steering analysis in Section~\ref{sec:steering_results}.

Introducing CAA and Direct Optimisation creates a comparison challenge. In particular, probe-based vectors arise from a linear classifier, whose held-out AUROC provides a direct measure of discriminative performance. CAA and Direct Optimisation however, do not, by construction, yield an equivalent classification score. We therefore define five direction stability metrics that allow the resulting directions to be compared without treating them as classifiers. Table~\ref{tab:geometry_metrics} summarises these metrics over the \tokenlyfont{\(\ell \times p \times m \times c\)} dimensions introduced in Section~\ref{sec:correctness_direction_axes}.

\begin{wrapfigure}[22]{r}{0.5\textwidth}
    \centering
    \vspace{-1.2em}
    \includegraphics[width=\linewidth]{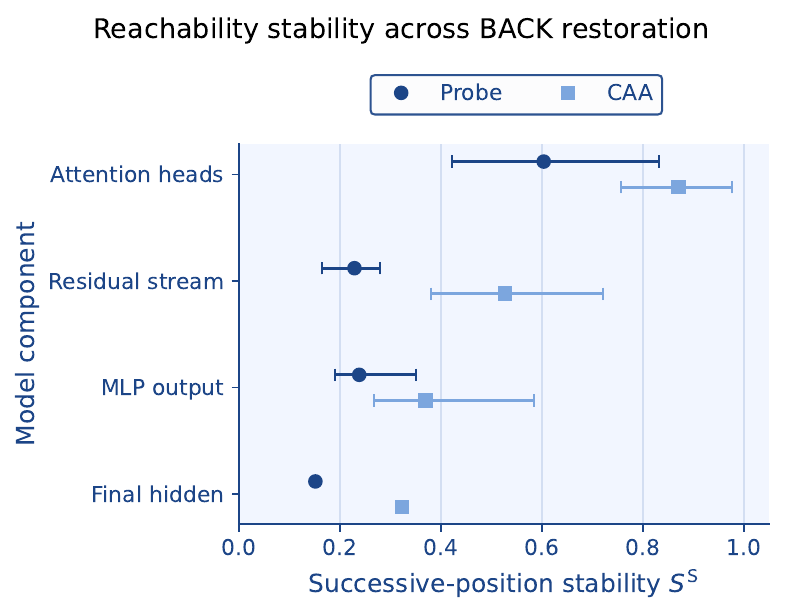}
    \vspace{-1.8em}
    \caption{Reachability directions remain more stable across \texttt{BACK} and state restoration at attention heads than at other model components. CAA has higher median stability than Probe in every component class. Points show median
    successive-position stability (\(S^{\mathrm{S}}\)), and horizontal bars show interquartile ranges.}
    \label{fig:successive_position_stability_back_reachability}
\end{wrapfigure}

\begin{figure}[!ht]
\centering
\includegraphics[width=1.0\linewidth]{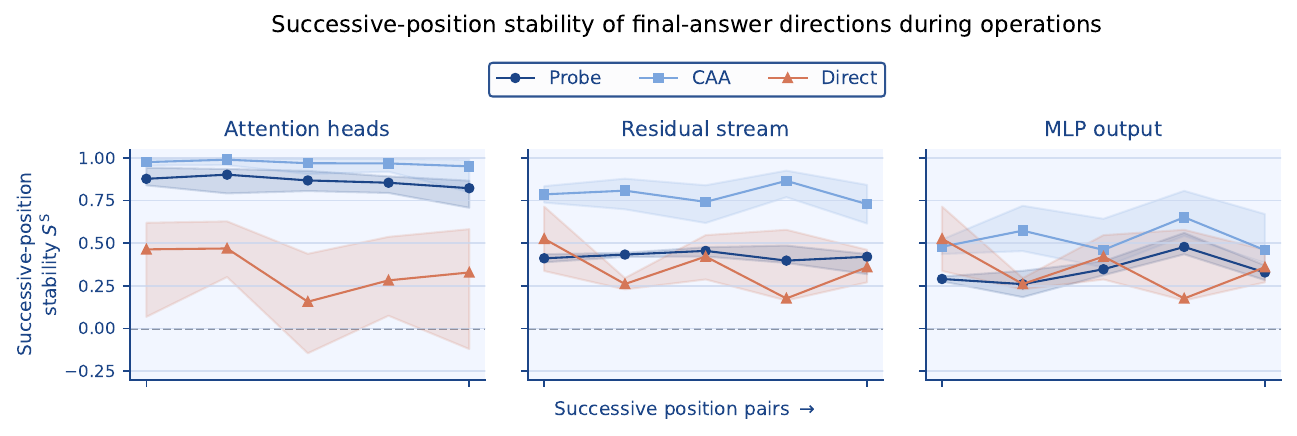}
\vspace{-1.5em}
\caption{Probe and CAA final-answer directions remain highly
stable across successive token positions within arithmetic state transitions at attention heads,
whereas Direct Optimisation yields weaker and more variable
alignment. Lines show median successive-position stability
(\(S^{\mathrm{S}}\)) within each component class, and shading
shows interquartile ranges. An expanded version is provided in
Appendix Figure~\ref{fig:successive_position_stability_operation_final_answer}.}
\vspace{0.5em}
\label{fig:successive_position_stability_operation_final_answer_wide}
\end{figure}

\begin{table}[!h]
\centering
\small
\caption{The five comparisons metrics A--E above, are used to test whether reasoning correctness forms a stable linear direction. Here, \(u_{m,p,c}\) denotes the unit-normalised direction obtained using method \(m\) at semantic position \(p\) and model component \(c\), \(\ell\in\{\mathrm{reach},\mathrm{answer}\}\) denotes the label family, and \(K\) is the number of selected semantic positions.}
\renewcommand{\arraystretch}{1.45}
\resizebox{\linewidth}{!}{
\begin{tabular}{
    @{\hspace{3ex}}
    >{\raggedright\arraybackslash}m{0.20\linewidth} 
    >{\centering\arraybackslash}m{0.34\linewidth} 
    >{\raggedright\arraybackslash}m{0.40\linewidth}
    @{}
}
\toprule
\textbf{Metric}
&
\textbf{Measure}
&
\textbf{Question}
\\
\midrule
\raisebox{1.0ex}{\textit{\makecell[l]{A. Successive-position\\\hspace{1.2em}stability}}}
&
\raisebox{0pt}{\(\displaystyle S^{\mathrm{S}}_{m,\ell,p\rightarrow p',c}=\cos\!\left(u^{\ell}_{m,p,c},u^{\ell}_{m,p',c}\right)\)}
&
What if we go one step at a time across successive reasoning positions?
\\
\midrule
\raisebox{1.0ex}{\textit{\makecell[l]{B. Cross-position \\\hspace{1.2em}stability}}}
&
\raisebox{0pt}{\makecell{\(\displaystyle S^{\mathrm{P}}_{m,c}=\frac{K\left\|\frac{1}{K}\sum_{p=1}^{K}u_{m,p,c}\right\|^2-1}{K-1}\)\\[-0.4ex]}}

&
\raisebox{1.0em}{\hspace{-0.0ex}\parbox{0.9\linewidth}{Does final-answer correctness share a common direction across reasoning positions?}}
\\
\midrule
\raisebox{1.0ex}{\textit{\makecell[l]{C. Cross-label\\\hspace{1.2em}stability}}}
&
\raisebox{0pt}{\(\displaystyle 
S^{\mathrm{L}}_{m,p,c}=\cos\!\left(u^{\mathrm{reach}}_{m,p,c},u^{\mathrm{answer}}_{m,p,c}\right)\)}
&
Do local reachability and eventual final-answer correctness share the same direction?
\\
\midrule
\raisebox{1.0ex}{\textit{\makecell[l]{D. Cross-method \\\hspace{1.2em}stability (Type 1)}}}
&
\raisebox{0pt}{
\(\displaystyle
S^{\mathrm{M1}}_{m_1,m_2,p,c}
=
\cos\!\left(
u_{m_1,p,c},
u_{m_2,p,c}
\right)
\)
}
&
Does the correctness direction remain stable across construction methods at the same position and component?
\\
\midrule
\raisebox{1.0ex}{\textit{\makecell[l]{E. Cross-method \\\hspace{1.2em}stability (Type 2)}}}
&
\raisebox{0pt}{
\(\displaystyle
S^{\mathrm{M2}}_{p,c}
=
\left\|
\operatorname{Proj}_{
\operatorname{span}
\{u^{P}_{p,c},u^{C}_{p,c}\}}
u^{D}_{p,c}
\right\|^2
\)
}
&
Does the Direct Optimisation vector lie within the plane spanned by Probe and CAA?
\\
\bottomrule
\end{tabular}
}
\label{tab:geometry_metrics}
\end{table}
A) \textit{Successive-position stability:}
Successive-position stability \(S^{\mathrm{S}}\) measures whether a direction persists between neighbouring semantic token positions in the reasoning trace. Using this metric, Figure~\ref{fig:successive_position_stability_back_reachability} shows that reachability directions change least from \texttt{BACK} to the restored state at attention heads, with CAA generally more stable than Probe. However, since this may simply reflect high certainty that \texttt{BACK} transitions tend to result in the same state (a reachable one), additional observations are required. Indeed,  Appendix Figure~\ref{fig:successive_position_stability_operation_reachability} shows a similar attention-head advantage during arithmetic state-transitions. Figure~\ref{fig:successive_position_stability_operation_final_answer_wide} applies the same analysis to final-answer directions, where Probe and CAA-based vectors again exhibit their greatest stability at attention heads.

\begin{figure}[!ht]
\vspace{-1.2em}
\centering
\includegraphics[width=0.95\linewidth]{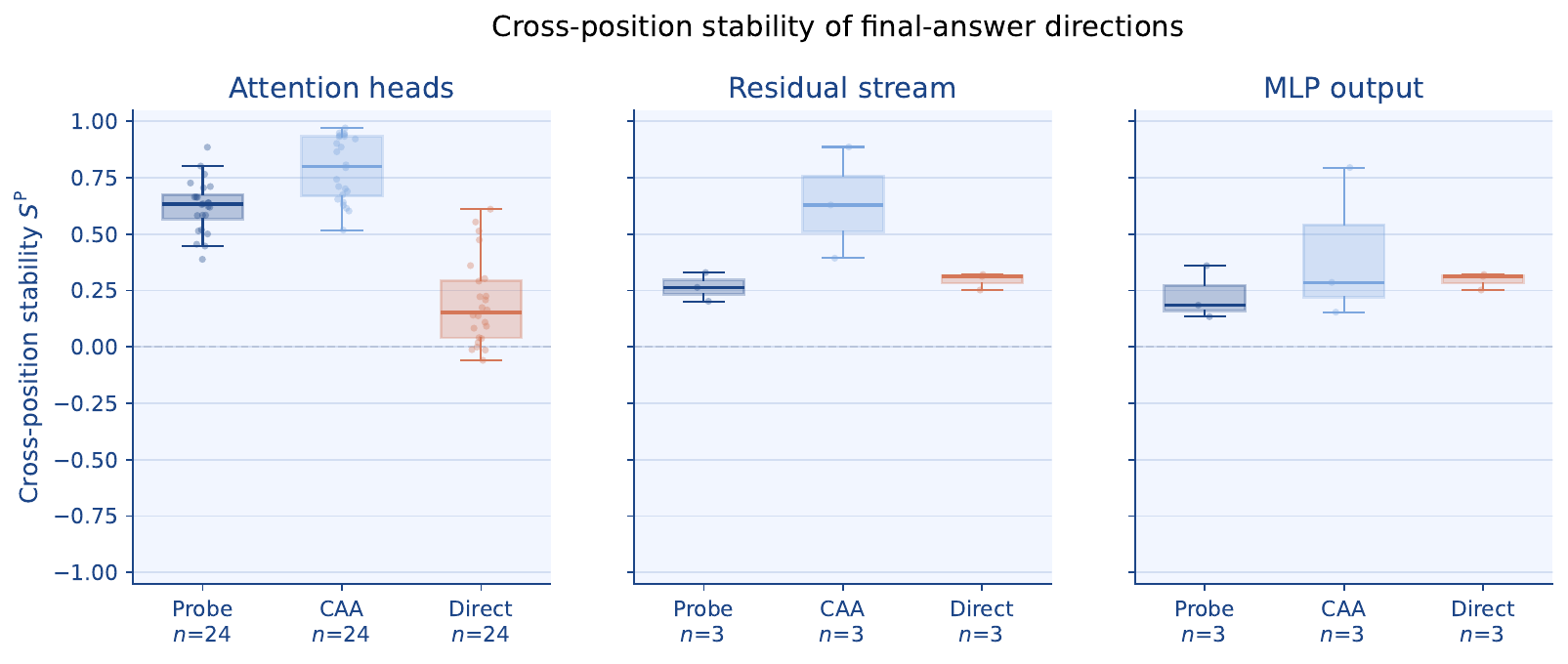}
\vspace{-1em}
\caption{We measure cross-position stability
(\(S^{\mathrm{P}}\)) to test whether final-answer
directions remain consistent across the selected
positions throughout the reasoning trace. CAA final-answer
directions have the highest median cross-position stability
in every component class. Probe directions are also more
stable across positions at attention heads than at other
components, whereas Direct Optimisation directions show
no comparable advantage. Each point reports stability
at one model location.}
\label{fig:cross_position_stability}
\vspace{-0.2em}
\end{figure}

B) \textit{Cross-position stability:} Cross-position stability \(S^{\mathrm{P}}\) asks whether final-answer correctness shares a common direction across the wider reasoning trace for a fixed construction method and model component. We report this measure only for final-answer correctness, since Direct Optimisation is not used together with reachability (see Section~\ref{sec:direct_optimisation_reachability}). Appendix Figure~\ref{fig:cross_position_stability} shows that CAA is most stable under this measure, with attention heads again exhibiting the greatest stability for the Probe and CAA methods.

C) \textit{Cross-label stability:}
Cross-label stability \(S^{\mathrm{L}}\) measures whether reachability and final-answer correctness share a direction at the same token position. Appendix Figure~\ref{fig:cross_label_stability_operation} shows that the attention-head ``advantage'' again extends from positional stability to agreement across correctness labels, particularly for CAA-based vectors. Even there, however, the cosine similarities remain moderate, indicating partial rather than complete agreement between the labels (a phenomenon addressed further in the concluding ``Observation'' to this subsection).
\clearpage
{\vspace{-2em}
D--E) \textit{Cross-method stability:}
Pairwise cross-method stability \(S^{\mathrm{M1}}\) tests whether directions constructed using different methods agree at the same token position and model component. Because pairwise cosine cannot determine whether the Direct Optimisation direction lies within the span of the Probe and CAA directions, the second type of cross-method stability \(S^{\mathrm{M2}}\) measures this as well. Figure~\ref{fig:direct_optimisation_probe_caa_span} combines these two metrics into a single plot. We may observe that Residual-stream and MLP directions remain close to the random projection baseline, whereas attention heads exhibit a broader distribution. However, high \(S^{\mathrm{M2}}\) scores are seen in a sparse subset of attention-head--position pairs (also interpreted in the concluding ``Observation'' below).}

\textbf{\underline{Observation}}: Across the five measures, attention heads most consistently exhibit the clearest directional stability. Indeed, the \textit{successive-position and cross-position stability} results could, on their own, suggest that final-answer correctness follows a common direction that persists throughout reasoning, particularly for Probe and CAA-based directions at those attention heads. \textit{Incomplete alignment across labels and methods does not inherently negate this token position stability observed within each method.}  Reachability and final-answer correctness are related but distinct measures of correctness, so the moderate alignment observed for CAA at attention heads, with cosine similarities around \(0.5\) (corresponding to angles of approximately \(60^\circ\)), is plausible. Nor should distinct construction methods necessarily recover identical vectors in a high-dimensional activation space. The broader spread at attention heads in Figure~\ref{fig:direct_optimisation_probe_caa_span} therefore remains notable. Primarily, these results further motivate the need to compare multiple correctness-direction methods during steering, rather than relying on a single method as some prior works may have opted to. Section~\ref{sec:steering_results} proceeds accordingly.

\section{Causal Steering}
\label{sec:steering_results} 
While Section~\ref{sec:experiments} examines whether correctness is linearly decodable and geometrically stable, this section asks whether the resulting directions can causally improve reasoning.

We evaluate steering across token position \(p\), correctness label \(\ell\), construction method \(m\), model component \(c\), and a new dimension \underline{intervention strength \(\beta\)}, yielding the search space \tokenlyfont{\(p \times \ell \times m \times c \times \beta\)}. The number of steered attention heads, and the intervention frequency, are also studied separately. Candidate intervention configurations are selected using validation data and evaluated on a held-out test split, with each steered generation paired with an unsteered generation of the same problem.

The two correctness label values \(\ell\) naturally allow us to examine causal effects both immediately after an intervention, and at the end of the generated trace:
\begin{enumerate}[leftmargin=2.0em, labelsep=0.6em, itemsep=0.45em, topsep=0.05em, parsep=0pt, partopsep=0pt] 

\item \textit{Next-state reachability} records whether the first state generated after an intervention can still reach the target using only valid forward arithmetic operations. For example, consider the \tokenlynogray{ChainSearch} problem ``\texttt{IN: [3,6,10,10,50,75]  TGT: 425}''. The legal move \texttt{6*75=450} produces \texttt{[3,10,10,50,450]}, which cannot reach the target without backtracking. By contrast, \texttt{10*50=500} produces \texttt{[3,6,10,75,500]}, from which \texttt{500-75=425} remains possible. We count a problem as improved if the next state is unreachable without steering but reachable with steering, and as worsened if the reverse occurs. 

\item \textit{Final-answer correctness} records whether steering changes the correctness of the completed answer. For the same problem, ``\texttt{IN: [3,6,10,10,50,75] TGT: 425}'', an unsteered response ending in \texttt{EXPR ((6*75)-10)} evaluates to \(440\) and is incorrect. If steering instead produces \texttt{EXPR ((10*50)-75)}, which evaluates to \(425\),
we count the problem as improved. If an originally correct
answer becomes incorrect under steering, we count it as worsened. Unlike next-state reachability, this measures whether the model actually reaches a correct answer, rather than whether a solution remains possible.
\end{enumerate}

Unlike probe reporting, causal steering requires fresh \textit{autoregressive generation} for every intervention condition. We therefore use a staged evaluation procedure in which 256 or 512 paired held-out problems provide an initial indication of whether an effect is statistically sufficient to justify evaluation across the full 16k-sample test split. 

We begin with attention heads, which Section~\ref{sec:experiments} identifies as the most favourable test case. Section~\ref{sec:attention_head_interventions} presents steering results for different construction methods across several attention-head counts, while Appendix Section~\ref{sec:additional_attention_head_interventions} tests whether these conclusions hold under different head-selection strategies, direction signs, and token positions, as well as whether steering with local-reachability directions affects final answers. Section~\ref{sec:steering_discussion} then broadens the analysis to model components beyond attention heads and compares different intervention frequencies. Appendix Section~\ref{sec:interventions_beyond_attention_heads} complements this discussion with additional experiments on intervention strength and vector normalisation, and documents how steering affects trace properties.

\subsection{Attention-Head Interventions}
\label{sec:attention_head_interventions}

\begin{figure}[!ht]
\centering
\vspace{-1em}
\includegraphics[width=1.0\linewidth]{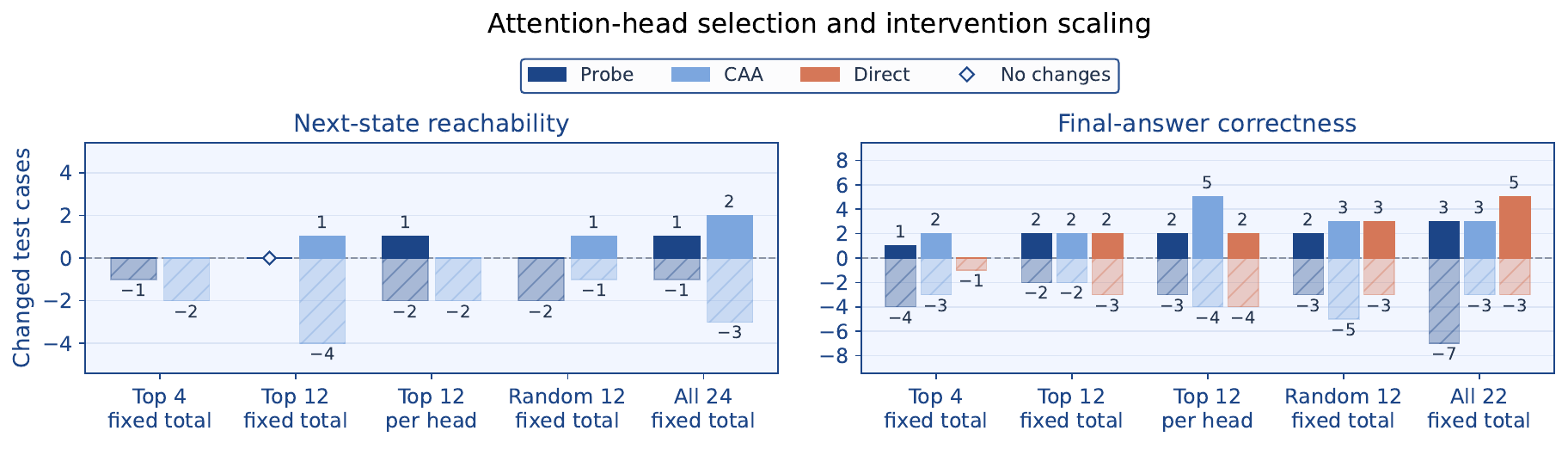}
\vspace{-1.5em}
\caption{Changing the number of steered attention heads yields no consistent improvement in next-state reachability or final-answer correctness. Each condition compares the same 256 test problems with and without steering. Solid upward bars count problems improved by steering,
hatched downward bars count problems worsened, and
diamonds indicate no change.}
\label{fig:head_set_and_scaling_controls_single_row}
\vspace{-0.5em}
\end{figure}

We first test whether correctness directions at the most promising attention heads improve reasoning on held-out problems. Figure~\ref{fig:head_set_and_scaling_controls_single_row} compares Probe, CAA, and Direct Optimisation across the highest-ranked head sets, and interventions over all available heads. We vary both the number of steered heads and whether intervention strength is fixed in total or applied separately to each head. Appendix Figure~\ref{fig:sparse_attention_head_heldout_counts} compares steering at attention heads selected using CAA separation with steering at randomly selected heads across several head counts. Neither selection strategy makes steering consistently improve next-state reachability or final-answer correctness.

\begin{figure}[!ht]
\centering
\vspace{0.5em}
\includegraphics[width=1.0\linewidth]{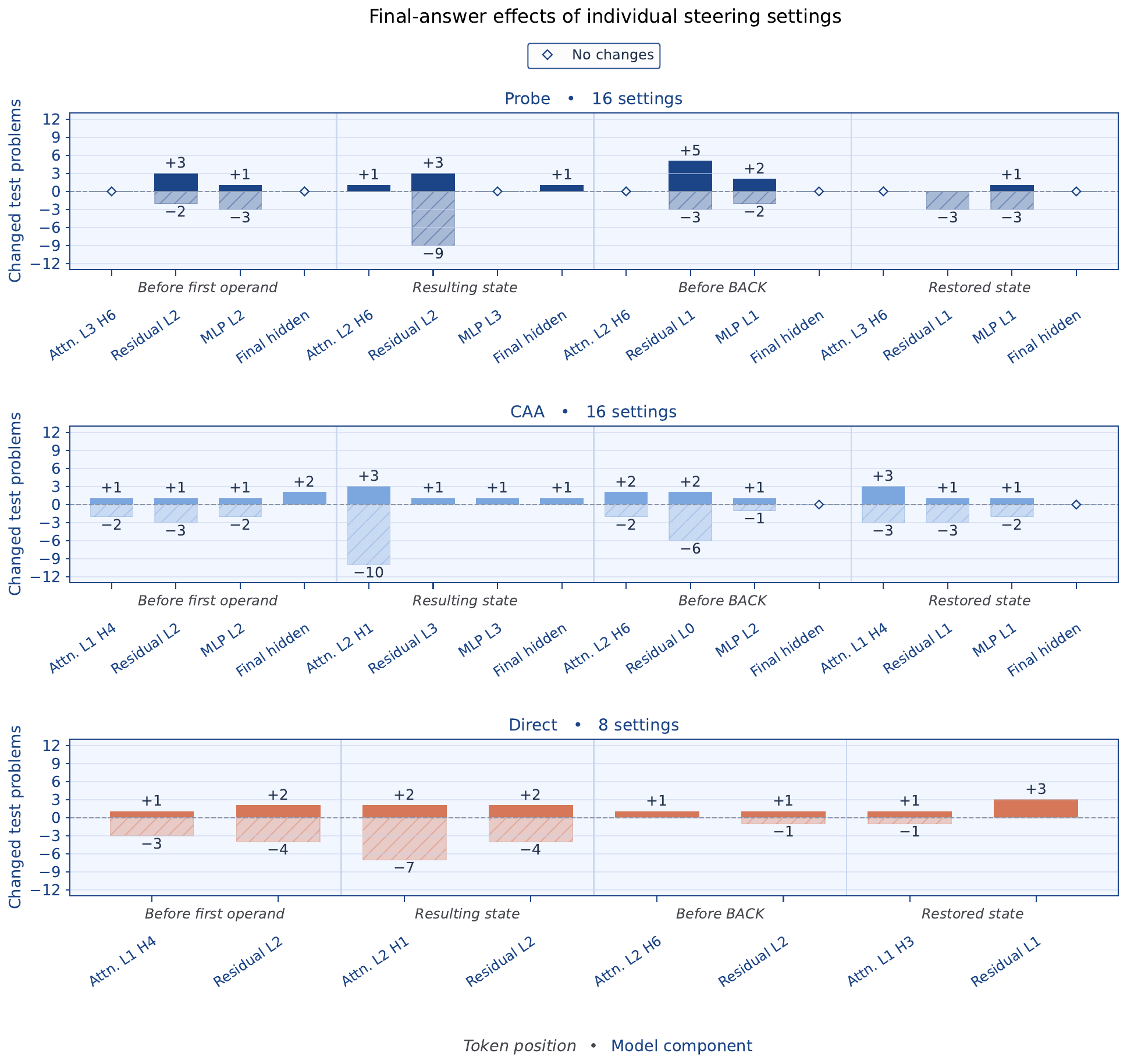}
\vspace{-2em}
\caption{Steering produces small, mixed changes in final-answer correctness, with several of the largest effects worsening outcomes. Upward bars count problems where the model originally gave an incorrect answer but gives a correct one after we apply steering and let it generate a new response. Downward bars count the opposite, where an originally correct answer becomes incorrect. Even Direct Optimisation, which explicitly trains a vector to increase the likelihood of a correct final answer (see Appendix~\ref{sec:direct_optimisation_reachability}), does not consistently improve correctness when the model generates a new response. Settings are chosen using validation data, then evaluated on 256 test problems, each with and without steering.}
\label{fig:steering_rescue_damage}
\vspace{-1.2em}
\end{figure}


Across the configurations in Figure~\ref{fig:head_set_and_scaling_controls_single_row}, steering changes individual outcomes in both directions without producing a consistent benefit. The largest displayed number of problems with improved final-answer correctness is 5 out of 256. CAA improves these 5 outcomes while worsening 4 others, and Direct Optimisation improves 5 while worsening 3. Probe steering
across all 22 attention heads instead improves 3 outcomes
and worsens 7. Changes in next-state reachability are similarly infrequent and occur in both directions. Appendix Figures~\ref{fig:heldout_position_local_global_counts} and~\ref{fig:attention_head_directionality_counts} likewise show that neither direction type nor direction sign consistently improves final-answer correctness across intervention positions.

In other words, varying which heads are steered, how many are steered, or the direction sign yields no dependable improvement in either next-state reachability or final-answer correctness.

\subsection{Broader Steering Interventions}
\label{sec:steering_discussion}

Building on the attention-head results in Section~\ref{sec:attention_head_interventions}, Figure~\ref{fig:steering_rescue_damage} compares Probe, CAA, and Direct Optimisation vectors across token positions and model components. Across these dimensions, steering moves some outcomes toward correctness and others away from it, without consistently improving held-out final-answer correctness.

\begin{figure}[H]
\centering
\vspace{-0.2em}
\includegraphics[width=1.0\linewidth]{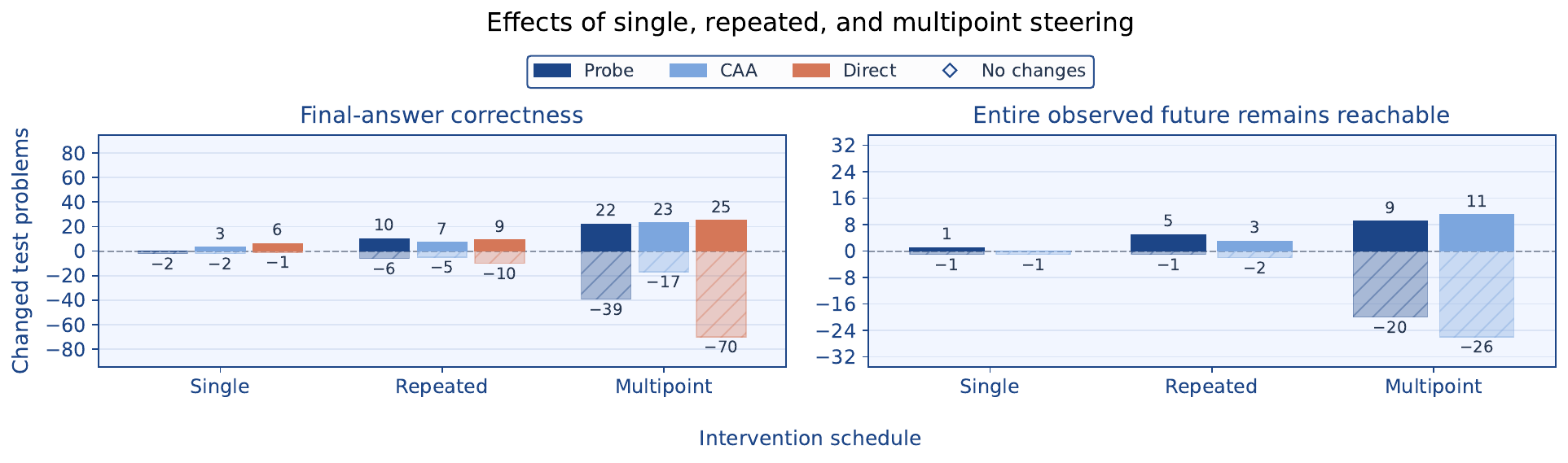}
\vspace{-1.9em}
\caption{Applying steering more frequently changes more outcomes, but does not consistently improve them. The largest increases in the number of worsened outcomes occur when steering is applied at multiple positions. Single steering acts before the first arithmetic state transition, repeated steering acts before every arithmetic state transition, and multipoint steering additionally acts after resulting and restored states, with settings chosen separately for each position. The panels measure final-answer correctness and whether every subsequent state remains reachable. Positive bars count problems improved by steering, while negative bars count problems worsened. Each setting is evaluated on 512 test problems, with and without steering.}
\label{fig:intervention_frequency_heldout_effects_compact_single_row}
\vspace{-1em}
\end{figure}

Figure~\ref{fig:intervention_frequency_heldout_effects_compact_single_row} shows that repeated and multipoint steering change more generations but mainly increase the number of problems with worsened outcomes. Appendix Figures~\ref{fig:where_and_when_dose_response} and~\ref{fig:steering_native_direct_pairs} extend this analysis, showing that neither varying intervention strength nor retaining the raw magnitude of Direct Optimisation vectors produces more reliable control.

Importantly, these opposing effects do not mean that the interventions are causally inert. Appendix Figure~\ref{fig:steering_side_effects} shows that steering changes generation length, \texttt{BACK} usage, and expression parseability without producing a common behavioural signature of improved reasoning.

\vspace{-0.2em}
\section{Conclusion}
\label{sec:conclusion}
\vspace{-0.2em}
We conduct a systematic study of reasoning correctness directions, asking whether linear decodability implies causal usefulness for steering. To do so, we introduce \tokenlynogray{ChainSearch}, a controlled arithmetic task in which reasoning traces can backtrack while searching for a solution. This simple setting allows us to analyse correctness directions across multiple dimensions and evaluate them through steering interventions. Correctness is linearly decodable, but our analysis reveals a nuanced picture of what constitutes a correctness direction. A direction that appears to capture correctness along one dimension may not do so along another. Even along dimensions where the recovered directions are largely stable, steering with them yields no consistent improvement. This suggests that familiar approaches to linear steering may not fully capture the representational structure relevant to improving reasoning accuracy, motivating new techniques focused on nonlinear representational geometries specific to \textit{reasoning correctness}.

\clearpage

\section*{AI Use Statement}
This research was developed by the authors through extensive discussion. We used AI primarily to assist with coding/debugging and to help polish the writing of the manuscript. The authors manually reviewed all work with great care and caution.

\section*{Reproducibility Statement}
We provide the code for dataset generation, model training, probing, and causal steering via the zip file in supplementary materials.


\bibliography{iclr2027_conference}
\bibliographystyle{iclr2027_conference}

\clearpage
\appendix
\section*{Appendix}
\section{Extended Related Work: Self-Correctness Signals}
\label{sec:related_work}

\begin{figure}[H]
\centering
\vspace{0.5em}
\includegraphics[width=1.0\linewidth]{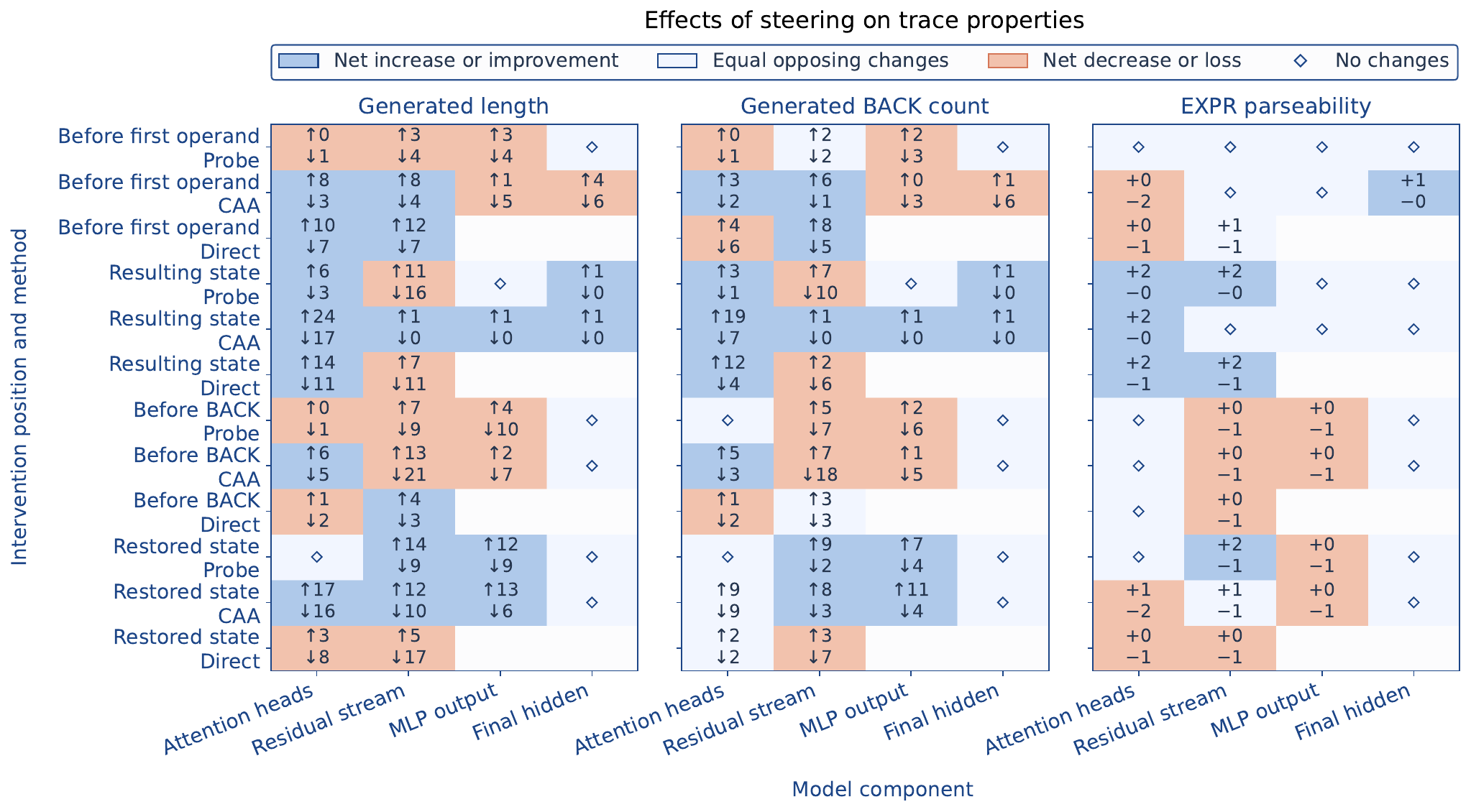}
\vspace{-1.8em}
\caption{We measure changes in generation length,
\texttt{BACK} usage, and expression parseability to examine
how steering affects the generated trace beyond final-answer
correctness. These effects vary across methods, model
components, and intervention positions, without a common
pattern of improved reasoning. The first two panels count
problems with increases and decreases separately, while
the third shows the number gaining a parseable expression
minus the number losing one. Each setting is evaluated
on 256 test problems, with and without steering.}
\label{fig:steering_side_effects}
\end{figure}

More broadly, developing methods to evaluate reasoning trajectories and understand the underlying reasoning processes of language models has attracted growing interest \citep{turpin2023language,chen2025reasoning,arcuschin2025chain,nwadike2026measuringaireasoningguide,nwadike2026einsteinworldmodels,shen2026understanding,shen2026detecting,shen2025faithcot}. With regards to steering, a key distinction from prior work is our focus on whether \textit{reasoning correctness itself} is associated with internal linear directions. \citet{venhoff2025understanding} instead steer behavioural patterns such as backtracking and explorativeness. However, \citet{chen2025seal} and \citet{zhuang2026reliable} show that regulating such patterns, including suppressing excessive reflection and transitions between approaches, can improve answer correctness. Yet, if linear steering improves correctness through behavioural changes, then a direction optimised directly for correctness should, in theory, recover both the improvement and the behavioural changes underlying it. Accordingly, we measure both aforementioned correctness changes, as well as changes in trace properties (see Figure~\ref{fig:steering_side_effects}).
 
 One study in this direction is by \citet{zhang2025reasoning}, who train linear probes to predict this intermediate answer correctness in a reasoning trace. A notable drawback of their setup is that it considers reasoning trace activations at the \textit{last model layer}. This last-layer focus is also seen in the work of \citet{ye2024physics}, who study GSM8K-style arithmetic traces with linear probes. However, the last model layer comes \textit{after} the model has completed all core computations relevant to its top-k token ranking, which leaves much of the model under-examined with regards to causal steering effects. 
 
 Indeed, \citet{zhang2025reasoning} avoid \textit{causal steering} altogether, and instead use their linear probes for early-exit, to decide when to halt token output. However, this experimental design decision leaves an evidence gap regarding their central question as to whether reasoning models know when they are right.

Another concern is that \citet{zhang2025reasoning} exclude unfinished generations from parts of their analysis. Their early-exit setup allows at most 10,000 tokens per problem. In our AIME 2025 reproduction with DeepSeek-R1-Distill-Qwen-7B, 56.7\% of reasoning traces exceeded this limit. This leaves the behavior of these unfinished cases unresolved.

\textit{Test-set contamination} is another concern. Because MATH was public before the evaluated models were trained, an early-exit probe may partly detect memorized answer patterns rather than online correctness. One way to rule out this concern completely would be to \textit{train the model from scratch} with full control over its training data. However, training a language model on a reasoning task from scratch is non-trivial, and is therefore an important component of the gap in the research literature that \tokenlynogray{ChainSearch} addresses.

Causal steering, in any case, may take on a number of definitions, which we discuss in greater detail in Appendix Section~\ref{sec:types_of_linear_steering}. In brief, this work uses probe-based, mean-difference, and directly optimized activation steering. For \textit{probe-based steering}, the steering vector is learned by minimizing a binary prediction loss, \(v^{*}=\arg\min_v \sum_i \mathcal{L}(y_i,v^\top h_i+b)\), and is then used directly as the intervention direction \citep{li2023inference,genadi-etal-2026-sycophancy}. We compare probe-based steering with \textit{mean-difference steering}, whereby we adapt contrastive activation addition (CAA) \citep{li2023inference,rimsky-etal-2024-steering} using a direction constructed by contrasting correct and incorrect traces \(v=\mathbb{E}_{i}[h_i\mid\mathrm{correct}]-\mathbb{E}_{i}[h_i\mid\mathrm{incorrect}]\). For completeness, we also consider \textit{directly optimized activation steering}, where the steering vector $v$ is obtained as
\(v^{*}=\arg\min_v \sum_i \mathcal{L}(M(h_i+v),y_i)\)
\citep{cao2024personalized,dunefsky2025oneshot}. Such direct optimization is far more \textit{computationally expensive} (and less common) than the other two methods, requiring repeated forward and backward passes through the model. \tokenlynogray{ChainSearch} makes it practical because our model has only 3 million parameters rather than billions or trillions.

To keep our steering findings relevant beyond the \tokenlynogray{ChainSearch} setting, we also restrict interventions to localized model components such as attention heads, residual streams, and MLPs \citep{li2023inference,chen2025persona,lee2024mechanistic,genadi-etal-2026-sycophancy}. Model-wide interventions, while feasible in our 3-million parameter setting, may be less extensible to larger models and datasets.



\clearpage

\vspace{1.5em}
\section{Dataset Details}
\label{dataset_details}

\subsection{ChainSearch Complexity}
\label{chainsearch_complexity}

As mentioned in Sections~\ref{sec:intro} and ~\ref{sec:task_formulation}, \tokenlynogray{ChainSearch} is based on the six-input version of Countdown, in which numbers must be combined to reach a target. The four-input version of the game, known as ``Game of 24'', has become a common testbed for language-model reasoning \citep{yao2023tree,stechly2025selfverification}. The six-input version used in \tokenlynogray{ChainSearch} has also recently been used as a reasoning benchmark \citep{ye2025beyond}. Unlike Game of 24, Countdown permits larger input sets and variable targets. Its state space grows rapidly with the number of inputs, and its general decision problem is NP-complete \citep{katz2025seemingly}. For six inputs, the state-space bound derived by \citet{katz2025seemingly} is over \(5{,}400\) times larger than for the four-input Game of 24 setting.

More generally, the search space grows exponentially with the number of inputs. The six-input version of Countdown used in \tokenlynogray{ChainSearch} contains approximately \(1.2\times10^7\) solvable instances~\citep{colton2014countdown}. For a fixed input such as \(\texttt{[3,6,10,10,50,75]}\), there are
\(6!\cdot\mathrm{Catalan}(5)\cdot4^5=30{,}965{,}760\) ways to order the inputs, group them with parentheses, and choose an operator for each calculation before removing equivalent constructions or invalid calculations. For example, \(\texttt{((75 * 6) - (50 / 10)) - (10 + 3)}\) and \(\texttt{((75 - 50) * (10 + 6)) + (10 * 3)}\) are two different fully parenthesised expressions constructed from the same six inputs. Thus, even a single compact problem permits millions of possible arithmetic constructions.

\tokenlynogray{ChainSearch} retains this six-input setting while representing search explicitly through state transitions and \texttt{BACK}. We use the HR3 system to enumerate all solvable instances \citep{colton2014countdown}, then generate supervision using search traces by solving each instance with a modified variant of A\textsuperscript{*} \citep{hart1968formal}, as explained in Appendix~\ref{sec:why_astar} below.




\subsection{Constructing Search Traces with Modified A*}
\label{sec:why_astar}
To construct the search traces used to fine-tune \tokenlynogray{ChainSearch}, we solve Countdown instances with a modified form of A* \citep{hart1968formal}. The HR3 solver described in Appendix~\ref{chainsearch_complexity} provides final expressions, whereas A* lets us construct structured traces showing arithmetic progress toward the target, unsuccessful attempts, and revisions via \texttt{BACK}.

\underline{Note}: A* is used only to construct the supervised finetuning (SFT) data. We \textit{do not} assume that the trained model internally implements A*, or that it explicitly stores alternative search states. The model learns to do reasoning \textit{entirely from a simple cross-entropy loss on the resulting SFT reasoning traces of how solve the input problems.}

A* is used to generate SFT reasoning traces using a cost function: \[f(n)=g(n)+h(n),\] where \(g(n)\) measures the cost already incurred and \(h(n)\) estimates the remaining cost of reaching the target. In our setting, \(g(n)\) counts the arithmetic operations already taken, while \(h(n)\) scores the current arithmetic state using the human-inspired strategies listed in Appendix~\ref{sec:heuristics_list}. Combined, these two terms encourage principled search traces in which the next branch is selected using both the path already taken and an intuitive estimate of the remaining distance to the target, rather than through brute-force exploration.

We make two \underline{\textit{modifications}} when applying A* as a tool for generating \tokenlynogray{ChainSearch} data:

First, conventional Countdown solutions record only the chosen arithmetic operations and do not explicitly represent abandoned attempts. We therefore \textit{linearise} the usual branching search of A* into a sequence of state transitions, inserting \texttt{BACK} whenever the procedure abandons an unproductive branch and returns to an earlier state.

Second, we do not require \(h(n)\) to be \textit{admissible}. In standard A*, admissibility means that \(h(n)\) never overestimates the true remaining cost, which supports a guarantee that the resulting path is optimal. Our objective is not to recover the shortest possible trace, but to generate informative search behaviour shaped by strategies used by human Countdown players. We therefore retain the \(g(n)+h(n)\) scoring framework while utilising the Countdown-specific heuristics of Appendix~\ref{sec:heuristics_list}.

\subsection{Dataset Splits}
\label{sec:dataset_splits}

\begin{SaveVerbatim}[commandchars=\\\{\}]{chaintrace_shared_target_backtracks}
\underline{A \protect\tokenlynogray{ChainSearch} Problem}
IN: [ 1, 2, 6, 8, 25, 50 ] TGT: 202

[ 1, 2, 6, 8, 25, 50 ] -> 8 * 25 = 200 -> [ 1, 2, 6, 50, 200 ] ;
BACK -> [ 1, 2, 6, 8, 25, 50 ] ;
[ 1, 2, 6, 8, 25, 50 ] -> 6 * 25 = 150 -> [ 1, 2, 8, 50, 150 ] ;
BACK -> [ 1, 2, 6, 8, 25, 50 ] ;
[ 1, 2, 6, 8, 25, 50 ] -> 6 * 50 = 300 -> [ 1, 2, 8, 25, 300 ] ;
BACK -> [ 1, 2, 6, 8, 25, 50 ] ;
[ 1, 2, 6, 8, 25, 50 ] -> 8 * 25 = 200 -> [ 1, 2, 6, 50, 200 ] ;
[ 1, 2, 6, 50, 200 ] -> 2 + 200 = 202 -> [ 1, 6, 50, 202 ] ;
SUCCESS ;
EXPR ((2 + (8 * 25))) \raisedcheckmark

\underline{A Different Problem with the Same Solution}
IN: [ 2, 3, 6, 8, 25, 50 ] TGT: 202

[ 2, 3, 6, 8, 25, 50 ] -> 8 * 25 = 200 -> [ 2, 3, 6, 50, 200 ] ;
BACK -> [ 2, 3, 6, 8, 25, 50 ] ;
[ 2, 3, 6, 8, 25, 50 ] -> 6 * 25 = 150 -> [ 2, 3, 8, 50, 150 ] ;
BACK -> [ 2, 3, 6, 8, 25, 50 ] ;
[ 2, 3, 6, 8, 25, 50 ] -> 6 * 50 = 300 -> [ 2, 3, 8, 25, 300 ] ;
BACK -> [ 2, 3, 6, 8, 25, 50 ] ;
[ 2, 3, 6, 8, 25, 50 ] -> 8 * 25 = 200 -> [ 2, 3, 6, 50, 200 ] ;
[ 2, 3, 6, 50, 200 ] -> 2 + 200 = 202 -> [ 3, 6, 50, 202 ] ;
SUCCESS ;
EXPR ((2 + (8 * 25))) \raisedcheckmark
\end{SaveVerbatim}

\begin{table}[!h]
\centering
\small
\caption{Allocation of \protect\tokenlynogray{ChainSearch} data across model pretraining, supervised fine-tuning, interpretability analysis, and causal steering. Split counts describe distinct traces or examples, whereas steering counts describe paired condition--problem evaluations. The same held-out problem may therefore be evaluated under several independently selected steering conditions.}
\resizebox{\linewidth}{!}
{\tablethreefont
\renewcommand{\arraystretch}{1.18}
\begin{tabular}{@{}
>{\RaggedRight\arraybackslash}p{0.30\linewidth}
@{\hspace{0.015\linewidth}}
>{\RaggedLeft\arraybackslash}p{0.23\linewidth}
@{\hspace{0.015\linewidth}}
>{\RaggedRight\arraybackslash}p{0.45\linewidth}
@{}}
\toprule
\textbf{Stage}
&
\textbf{Size}
&
\textbf{Purpose}
\\
\midrule

\multicolumn{3}{@{}l}{\textit{Model pretraining}} \\
\addlinespace[0.25em]

Exploratory pretraining corpus
& 6,948,231 traces
& Procedurally generated search with controlled backtracking \\

\quad Pretraining train
& 6,947,727 traces
& Model pretraining \\

\quad Pretraining validation
& 504 traces
& Pretraining checkpoint selection \\

\midrule

\multicolumn{3}{@{}l}{\textit{Supervised fine-tuning and held-out allocation}} \\
\addlinespace[0.25em]

Complete \tokenlynogray{ChainSearch} corpus
& 1,016,676 examples
& Supervised fine-tuning and held-out analyses \\

\quad Fine-tuning train
& 808,033 examples
& Model training \\

\quad Fine-tuning validation
& 900 examples
& Checkpoint selection \\

\quad Fine-tuning test
& 9,000 examples
& Final model evaluation \\

\quad Held-out pool for correctness-\newline\hspace{1em}direction analysis
& 115,367 examples
& Set aside before fine-tuning for correctness-direction analysis \\

\quad Unassigned held-out reserve
& 83,376 examples
& Reserved for target-disjoint splitting (see Appendix Section~\ref{sec:model_training}) \\

\midrule

\multicolumn{3}{@{}l}{\textit{Correctness direction analysis}} \\
\addlinespace[0.25em]

Generated traces for correctness-direction analysis
& 115,367 traces
& Shared source for Probe, CAA, Direct Optimisation, and evaluation \\

\quad Train split
& 83,354 traces
& Correctness direction construction \\

\quad Validation split
& 15,899 traces
& Candidate and hyperparameter selection \\

\quad Test split
& 16,114 traces
& Held-out probe reporting and steering source pool \\

\addlinespace[0.15em]

Labelled token position instances
& 6,877,165 instances
& Position-level correctness supervision \\

Cached activation positions
& 6,335,310 positions
& Probe fitting and CAA mean-difference construction \\

\midrule

\multicolumn{3}{@{}l}{\textit{Causal steering evaluation}} \\
\addlinespace[0.25em]

Steering test per condition
& 256 or 512 problems
& Initial paired causal evaluation \\

Complete held-out steering suite
& 68,096 paired evaluations
& 236 selected conditions across four experiments \\

\bottomrule
\end{tabular}%
}
\label{tab:dataset_allocation}
\end{table}

Table~\ref{tab:dataset_allocation} summarises how data is allocated across model training, correctness direction evaluation, and causal steering.

Probe results are reported on the full 16,114-trace test split. Steering evaluations instead use 256 or 512 problems per selected condition, because each condition requires fresh autoregressive generation. These subsets provide an initial test of whether an intervention produces effects large and consistent enough to warrant evaluation on the full test split.

Example~\ref{example:shared_target_backtracking} illustrates why target-disjoint splitting is necessary, as problems sharing a target can produce nearly identical reasoning traces even when their inputs differ slightly. Further details on the splitting procedure are provided in Appendix Section~\ref{sec:model_training}. 

Because targets are kept disjoint across splits, examples must be allocated in target-level groups. Constructing the fine-tuning, evaluation, and interpretability subsets at the designated sizes therefore leaves some examples unassigned. We retain these examples as the ``Unassigned held-out reserve'' reported in Table~\ref{tab:dataset_allocation}. 

Appendix~\ref{sec:additional_trace_examples} provides more example of \tokenlynogray{ChainSearch} traces containing one to eight backtracks. 

\vspace{1em}
\begin{chainexamplebreakable}[label={example:shared_target_backtracking}]{Nearby inputs can share the same three-backtrack solution structure}
\UseVerbatim[fontsize=\chaintracefontsize]{chaintrace_shared_target_backtracks}
\end{chainexamplebreakable}

\vspace{3em}
\section{Additional Probing Plot Result}
\label{sec:appendix_extra_probing_results}
\begin{figure}[!h]
\centering
\vspace{0em}
\includegraphics[width=0.5\linewidth]{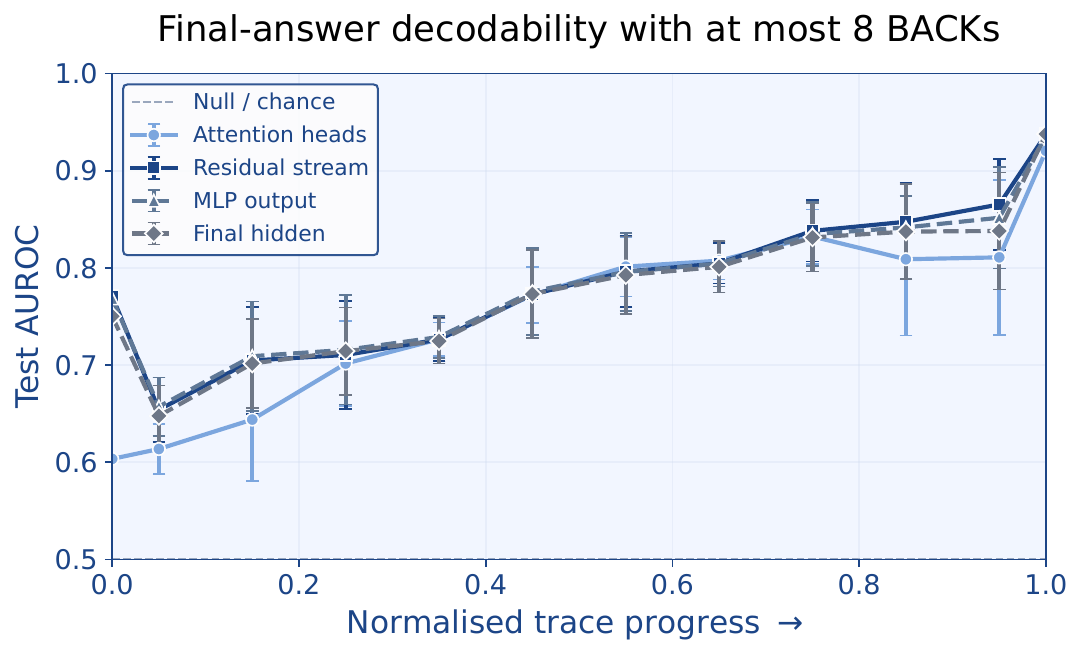}
\caption{As in Figure~\ref{fig:final_answer_decodability_trace_progress}, final-answer correctness becomes more decodable as generation progresses, even when we restrict traces to at most eight \texttt{BACK} transitions. This suggests
that the increase is not primarily driven by traces with unusually frequent backtracking. Error bars show one standard deviation around the mean.}
\label{fig:final_answer_decodability_trace_progress_max_8_backtracks}
\end{figure}

\vspace{2.5em}
\clearpage
\section{Model Training}
\label{sec:model_training}

We train the model using \textit{standard token-level cross-entropy} over the trace, with the problem specification masked from the loss. The trajectories themselves provide supervision for arithmetic and search behaviour (without additional losses that explicitly reward correct operations or other desired behaviours). 

We use digit-wise tokenisation, representing \texttt{899} as \texttt{8 9 9}, so that a fixed vocabulary can express the larger values produced by intermediate calculations. For correctness direction analysis, the last token position within a given number is considered as representative of the entire number when varying the semantic token position \(p\) (introduced in Section~\ref{sec:correctness_direction_axes}). 

In preliminary experiments, we trained the model directly on traces generated using these heuristics within a modified A\textsuperscript{*} algorithm (see Appendix~\ref{sec:why_astar}). However, the model's generated traces contained arithmetic errors, inconsistent states, and repetitive arithmetic operations despite stable training loss. We therefore include a pretraining phase, where the model is exposed to randomly generated exploratory traces containing valid arithmetic operations, state transitions, and backtracking, without requiring that they reach the target. This stage is intended to familiarise the model with valid trace structure before subsequent fine-tuning on A\textsuperscript{*} solution traces.

\begin{SaveVerbatim}[commandchars=\\\{\}]{chaintrace_target_101}
\underline{Problem}
IN: [ 1, 8, 8, 9, 9, 100 ] TGT: 101

\underline{Solution}
[ 1, 8, 8, 9, 9, 100 ] -> 100 + 1 = 101
-> [ 8, 8, 9, 9, 101 ] ;
SUCCESS ;
EXPR ((100 + 1)) \raisedcheckmark
\end{SaveVerbatim}

\textit{How do we separate training and evaluation when
distinct problems can share the same solution?}

\begin{wraptable}[12]{r}{0.48\textwidth}
\centering
\small
\vspace{-1.0em}
\parbox{\linewidth}{%
\caption{Target-disjoint splitting is motivated by the observation that even consecutive targets can lead to substantially different solution structures for a fixed input, here \texttt{IN: [1,8,8,9,9,100]}.}
\vspace{-0.8em}
\label{tab:adjacent_target_solutions}}
\renewcommand{\arraystretch}{1.15}
\setlength{\tabcolsep}{4pt}
\begin{tabular}{@{}clc@{}}
\toprule
\textbf{Target} & \textbf{Final \texttt{EXPR}} & \textbf{Steps} \\
\midrule
101 & \texttt{(100+1)}                 & 1 \\
102 & \texttt{(100+(9-(8-1)))}         & 3 \\
103 & \texttt{(100+(9-((8-(9-8))-1)))} & 5 \\
104 & \texttt{(100+(8/(9-(8-1))))}     & 4 \\
\bottomrule
\end{tabular}
\end{wraptable}

Because the task does not require using every input, distinct problems can share the same solution and search structure. For example, the inputs ``\texttt{IN: [1,2,6,8,25,50] TGT: 202}'' and  ``\texttt{IN: [2,3,6,8,25,50] TGT: 202}'', both admit the same solution of \texttt{EXPR (2+(8*25))}. Furthermore, their corresponding reasoning traces would also be highly similar (besides the single number swap) as illustrated in Example~\ref{example:shared_target_backtracking} in Appendix~\ref{sec:dataset_splits}, requiring a similar sequence of arithmetic operations and \texttt{BACK} transitions.

\textbf{Target-Disjoint Splitting}: We address this by assigning all problems with the same target to the same split, reducing the risk that closely related
solution traces appear in both training and evaluation.
Changing the target, even between consecutive integers, can lead to substantially different solutions while the inputs remain fixed (see Table~\ref{tab:adjacent_target_solutions}).

\textbf{Model Competence}: Table~\ref{tab:model_competence} reports how reliably our model generates valid arithmetic operations, consistent state transitions, and correct final expressions after each training stage.

\begin{table}[!ht]
\centering
\small
\caption{The model achieves high arithmetic operation validity and state
transition consistency, suggesting its competence.  A final-answer accuracy of (71\%) provides a useful middle ground where performance is strong, while the remaining (29\%) of traces provide enough negative examples to limit class imbalance during probe training and CAA construction. Pretraining results use 504 validation
examples, and fine-tuning results use 9,000 test examples (matching the corresponding data splits outlined in Appendix~\ref{sec:dataset_splits} Table~\ref{tab:dataset_allocation}).}
\label{tab:model_competence}
\renewcommand{\arraystretch}{1.15}
\setlength{\tabcolsep}{5pt}
\begin{tabular*}{\linewidth}{@{\extracolsep{\fill}}p{0.22\linewidth}p{0.43\linewidth}rr@{}}
\toprule
\raisebox{1.0ex}{\textit{Metric}}& \raisebox{1.0ex}{\textit{What it measures}}
& \shortstack{\textit{After}\\\colorbox{pretrainpastel}{\textit{pretraining}}}
& \shortstack{\textit{After}\\\colorbox{finetunepastel}{\textit{fine-tuning}}} \\
\midrule
Expression parseability
& The final \texttt{EXPR} can be parsed
& 95.63\% & 99.33\% \\

Arithmetic operation validity
& A generated arithmetic operation is legal
& 93.58\% & 96.37\% \\

State consistency
& The resulting state from a state transition correctly reflects the arithmetic operation
& 92.93\% & 95.16\% \\ 

\hspace{-4.25em}\makecell{Final-answer \\ correctness}
& The final \texttt{EXPR} evaluates to the target

& 25.20\% & 71.12\% \\
\bottomrule
\end{tabular*}
\end{table}

\clearpage 


\clearpage
\section{Types of Linear Steering}
\label{sec:types_of_linear_steering}

We differ from prior work by focusing specifically on whether reasoning correctness itself can be steered. \citet{venhoff2025understanding} instead steer higher-level reasoning behaviours such as uncertainty, example testing, and backtracking. \citet{chen2025seal} and \citet{zhuang2026reliable} show that steering behaviours such as reflection and transition can indirectly improve answer accuracy, but the underlying directions encode those behaviours rather than correctness itself. Four constructions are especially relevant, as summarized in Table~\ref{tab:linear_direction_constructions}.
\begin{table*}[!h]
\centering
\small
\caption{
Four constructions for linear intervention directions. Here, \(h_{i,p,c}\) denotes the activation for example \(i\) at token position \(p\) and model component \(c\), \(y_i\in\{0,1\}\) denotes the corresponding binary label, and \(\theta_{ft}\) and \(\theta_{pre}\) denote the parameters of the finetuned and pretrained model respectively. Here, \(M\) denotes the frozen model, \(v\) denotes the direction produced by each construction, \(b\) denotes the learned probe bias, with its joint optimization omitted from the notation for brevity, \(a_i\) denotes the target answer-token sequence for example \(i\), and \(\mathcal{L}\) denotes the classification loss for the probe, and the objective for direct optimisation direction, respectively.
}
\renewcommand{\arraystretch}{1.6}
\resizebox{\textwidth}{!}{
\begin{tabular}{
    @{}
    >{\raggedright\arraybackslash}p{0.12\textwidth}|
    >{\centering\arraybackslash}m{0.16\textwidth}|
    >{\centering\arraybackslash}m{0.21\textwidth}|
    >{\centering\arraybackslash}m{0.25\textwidth}|
    >{\centering\arraybackslash}m{0.25\textwidth}
    @{}
}
\toprule
&
\textbf{Parameter-space task vector}
&
\textbf{Probe direction}
&
\textbf{Mean-difference activation direction (CAA)}
&
\textbf{Directly optimized activation direction}
\\
\midrule

\raisebox{-1.4ex}{
\hspace{-.8ex}\textbf{Construction}
}
&
\raisebox{-1.4ex}{
\(\displaystyle
v=\theta_{ft}-\theta_{pre}
\)
}
&
\(\displaystyle
\begin{gathered}
v_{p,c}
=
\arg\min_v\sum_i\{\\
\mathcal{L}(y_i,v^\top h_{i,p,c}+b)\}
\end{gathered}
\)
&
\raisebox{-1.4ex}{
\(\displaystyle
\begin{gathered}
v_{p,c} = \mathbb{E}\!\left[h_{i,p,c}\mid y_i=1\right] - \\
\mathbb{E}\!\left[h_{i,p,c}\mid y_i=0\right]
\end{gathered}
\)
}
&
\(\displaystyle
\begin{gathered}
v_{p,c}
=
\arg\min_v \sum_i\{\\
\mathcal{L}\left(M(h_{i,p,c}+v),a_i\right)\}
\end{gathered}
\)
\\
\midrule
\textbf{Space}
&
Parameter space
&
Activation space
&
Activation space
&
Activation space
\\

\bottomrule
\end{tabular}
}
\label{tab:linear_direction_constructions}
\end{table*}

Our causal experiments compare the three activation-space direction vector methods, but the parameter-space task vector method is a nice contrast:

\begin{enumerate}[label={\Roman*.},leftmargin=*]
\vspace{-1em}
\item \label{item:task-vector}
\textit{Parameter-space task vectors}: Task vectors measure the parameter difference between two models, typically reflecting the changes introduced by task adaptation \citep{ilharco2023editing}. \textit{We do not use this approach}, and instead focus on the activation-space constructions in Items~\ref{item:probe},~\ref{item:mean-difference}, and~\ref{item:optimized}. This is because the parameter difference is not conditioned on correctness and may therefore capture broader task-related changes rather than correctness specifically. In particular, the direction vector \(v=\theta_{ft}-\theta_{pre}\) is not defined at a specific token position \(p\) or model component \(c\), and is therefore not denoted \(v_{p,c}\) like the activation-space methods.

\item \label{item:probe}
\textit{Probe-based steering}: A linear probe learns a discriminative direction that separates activations associated with the two classes. The learned probing weight vector $v_{p,c}$ can then be used directly as an intervention direction, as in prior work on properties such as truthfulness and sycophancy \citep{li2023inference,genadi-etal-2026-sycophancy}.

\item \label{item:mean-difference}
\textit{Mean-difference activation steering}: Contrastive Activation Addition constructs the steering direction $v_{p,c}$ from the displacement between the mean activations of two classes \citep{rimsky-etal-2024-steering}. Related approaches have been used to steer truthfulness and other high-level behaviours \citep{li2023inference,rimsky-etal-2024-steering}, as well as reasoning behaviours including uncertainty, backtracking, reflection, and transition \citep{venhoff2025understanding,chen2025seal,zhuang2026reliable}. We compute these directions at the same semantic positions and model components as the linear probes. This gives a direct comparison between a discriminatively learned direction (in II above) and the corresponding difference between class means.

\item \label{item:optimized}
\textit{Directly optimized activation steering}: Unlike probe-based and mean-difference steering, this approach does not derive its direction from a classifier or class-conditional activation statistics. Instead, it backpropagates through the frozen model while updating only an additive activation-space vector $v_{p,c}$ \citep{cao2024personalized,dunefsky2025oneshot}. In our setting, a reusable vector is optimized over many training examples and their target answers \(a_i\), for a given semantic position and model component. This provides a control for distinguishing failures of direction construction from broader limits of local additive steering.

\end{enumerate}

\clearpage
 
\section{Direct Steering Vector Optimisation Details}
\label{sec:direct_optimisation_reachability}

As discussed in Section~\ref{sec:correctness_direction_axes} and Appendix Section~\ref{sec:types_of_linear_steering}, Direct Optimisation provides an optimisation-based control experiment for comparison with probe- and CAA-based steering.  

\subsection{Direct Steering Vector Optimisation Objective}
\label{sec:direct_optimisation_objective}
\vspace{-0.5em}
For Direct Optimisation, we freeze the model \(M\) and optimise only the additive vector,
\[
v_{p,c}^{\mathrm{opt}}
=
\arg\min_v
\sum_i
\mathcal{L}_{\mathrm{ans}}
\left(
M(h_{i,p,c}+v),
a_i
\right),
\]
where \(a_i\) is the selected valid final-answer token sequence. \textit{When the model's generated expression is correct, we use that expression as \(a_i\). Otherwise, we use the corresponding A\textsuperscript{*} reference expression.} This avoids penalising valid expressions that differ syntactically from the reference.

The loss is computed on \(a_i\) conditioned on the model's original generated prefix,
\[
\mathcal{L}_{\mathrm{ans}}
\left(
M(h_{i,p,c}+v),
a_i
\right)
=
-\frac{1}{|a_i|}
\sum_{t=1}^{|a_i|}
\log
p_M
\left(
a_{i,t}
\mid
a_{i,<t},
h_{i,p,c}+v
\right),
\]
where the fixed preceding prefix is omitted from the conditioning notation. Thus, optimising \(v\) maximises the mean log-probability assigned to the selected valid final-answer tokens. We select the vector with the lowest loss on final-answer tokens in the validation data, then evaluate whether it improves final-answer correctness when the model generates a new response in the later steering experiments. Where multiple selected token positions from the same trace can contribute, examples are weighted so that each trace contributes equally, and we apply a small \(L_2\) penalty to \(v\).

\subsection{Optimisation Convergence}
\vspace{-0.5em}
During Direct Optimisation, we evaluate the vector every 20 optimisation steps. At each validation point \(t\), we measure the cosine similarity between the current vector and the vector from the preceding validation point, \(s_t=\cos\!\left(v_t,v_{t-20}\right).\) We treat a run as directionally stable when both the final value of \(s_t\) and its median over the final four validation intervals, corresponding to the last 80 optimisation steps, are at least \(0.999\). Values close to \(1\) indicate that the vector's orientation has stabilised even if its norm or validation loss continues to change.

\subsection{Supervision Signal}
\vspace{-0.5em}
\textit{Reachability} has no equally direct objective because \(M\) never predicts whether a state is reachable. One option is to introduce a reachability classifier, but then Direct Optimisation would depend on a probe. Another option is to let \(v\) affect subsequent generation and supervise whether the resulting states remain reachable. Because reachability can change at every step, this becomes repeated supervision over the evolving trace rather than optimisation of a single local reachability direction.

We therefore use final-answer supervision for Direct Optimisation, while the probe and CAA constructions retain both reachability and final-answer labels. \tokenlyfont{SUCCESS} remains eligible because \tokenlyfont{EXPR} follows it, while the final parenthesis of \tokenlyfont{EXPR} is excluded because the answer has already been produced.

\subsection{Gradient-Path Constraints}
\vspace{-0.5em}
Backpropagation imposes a further constraint because Direct Optimisation requires a differentiable path from \(v\) to the logits for later answer tokens. In our four-layer model, interventions at layers 0--2 can still influence later tokens through subsequent attention layers, so \(\partial \mathcal{L}_{\mathrm{ans}}/\partial v \neq 0\). At layer 3 and the final hidden representation, no subsequent attention layer remains to carry the perturbation forward, giving \(\partial \mathcal{L}_{\mathrm{ans}}/\partial v = 0\). We therefore exclude these locations from Direct Optimisation, leaving 30 eligible model locations per semantic position rather than the 41 used for probe and CAA vectors. Probe and CAA vectors from the excluded locations are retained because they can still affect free generation by changing the next emitted token, even though this discrete route cannot carry  gradients.


\section{Additional Correctness Direction Results}
\label{sec:additional_correctness_diection_interp}

\vspace{-0.2em}
\begin{figure}[H]
\centering
\includegraphics[width=1.0\linewidth]{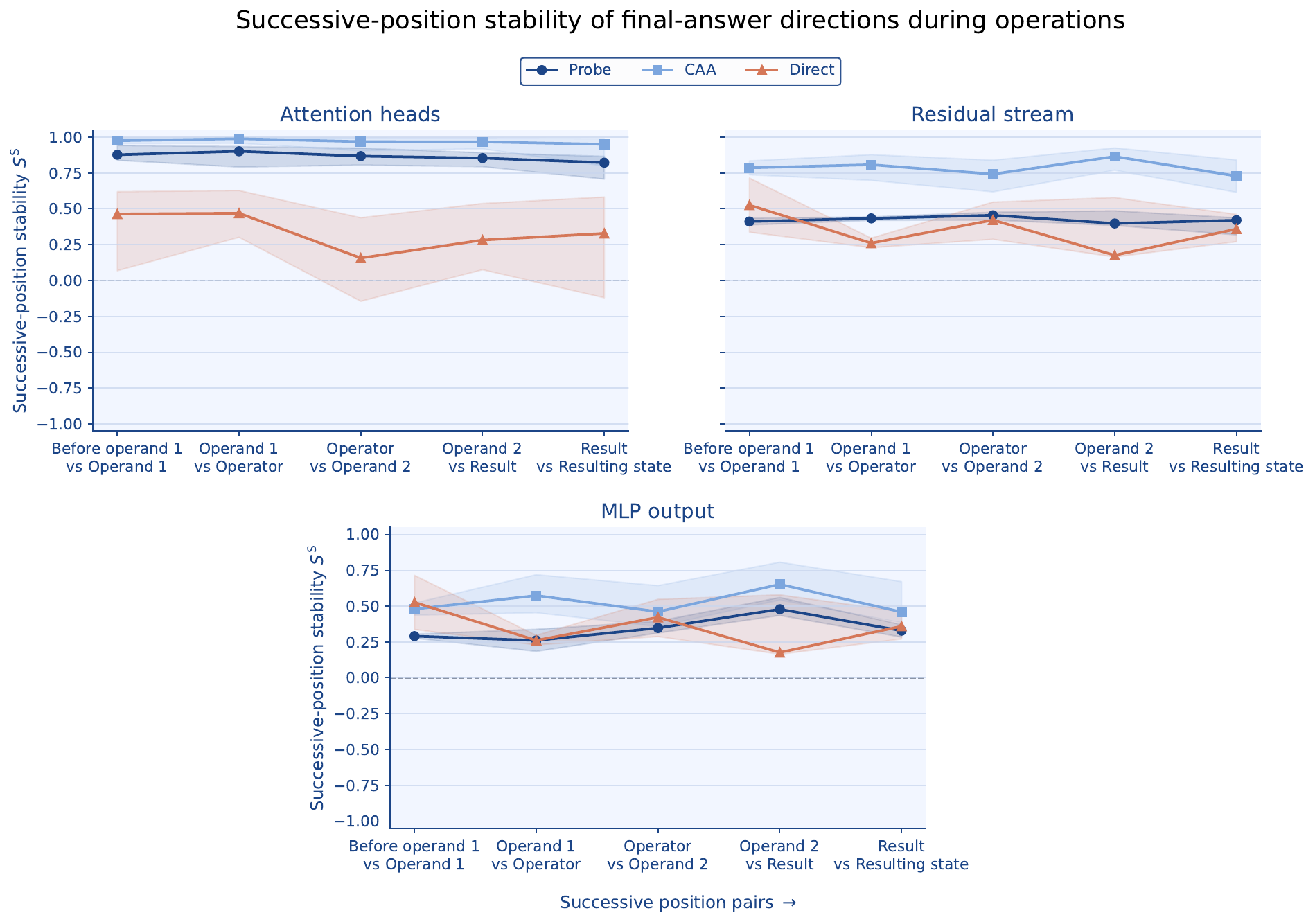}
\vspace{-1.5em}
\caption{We measure successive-position stability
(\(S^{\mathrm{S}}\)) to examine whether \textbf{\underline{final-answer}} directions remain similar as the model constructs an
arithmetic state transition. Probe and CAA final-answer directions
show high stability at attention heads, remaining similar
across successive token positions. Directions obtained
through Direct Optimisation show lower stability, while
residual-stream and MLP directions are less stable or
vary more in stability across positions. Lines show
medians across locations within each component class,
and shading shows interquartile ranges.}
\label{fig:successive_position_stability_operation_final_answer}
\end{figure}

\begin{figure}[H]
\centering
\vspace{-0.9em}
\includegraphics[width=1.0\linewidth]{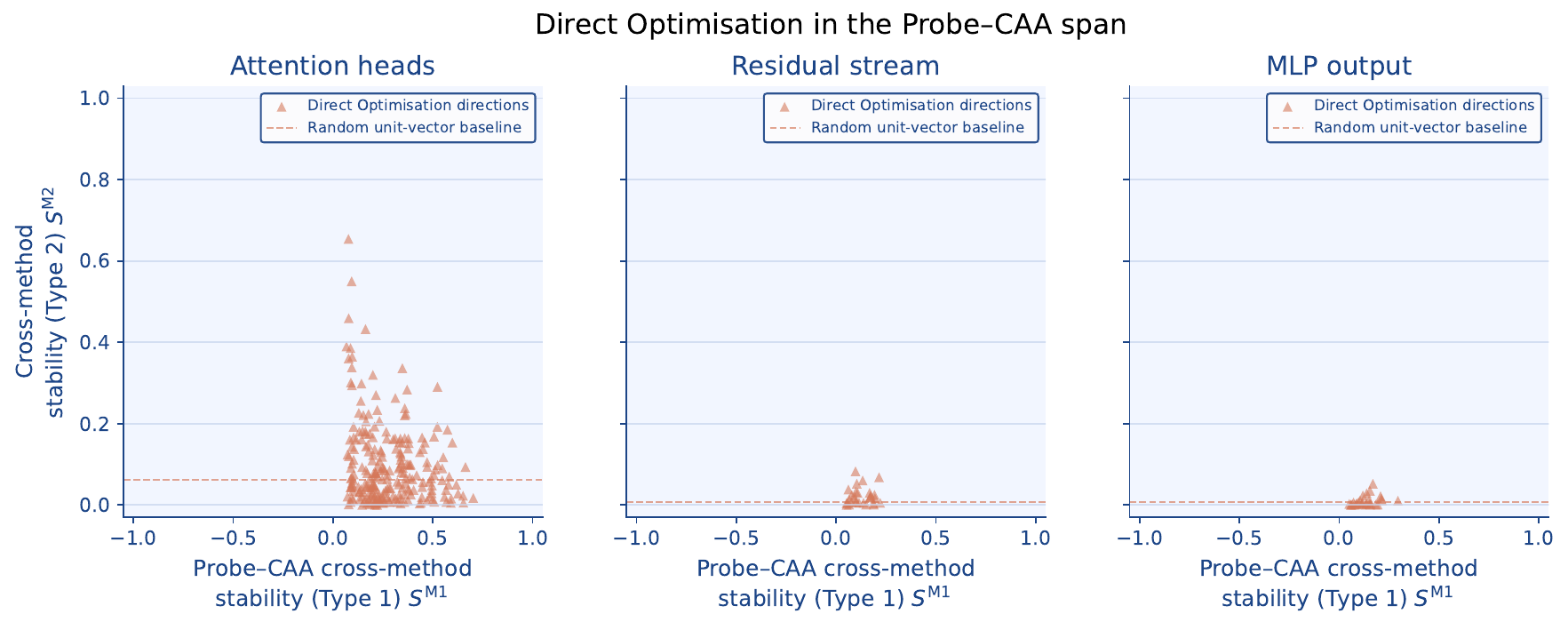}
\vspace{-2em}
\caption{Plotting cross-method stability Type 2
(\(S^{\mathrm{M2}}\)) against Probe--CAA cross-method
stability Type 1 (\(S^{\mathrm{M1}}\)) reveals that
the strongest agreement with Direct Optimisation is
concentrated at a small subset of attention-head locations.
\(S^{\mathrm{M2}}\) measures how much of a Direct
Optimisation direction lies within the plane spanned
by Probe and CAA, even when its cosine similarity with
either vector individually is low. Some attention-head
directions have substantial projections onto this plane,
while projections at residual streams and MLP outputs
remain close to the random baseline. Each point compares
directions at the same token position and model location.
Dashed lines show median projections of random unit vectors.}
\label{fig:direct_optimisation_probe_caa_span}
\vspace{-1em}
\end{figure}

\clearpage

\begin{figure}[H]
\centering
\vspace{1em}
\includegraphics[width=1.0\linewidth]{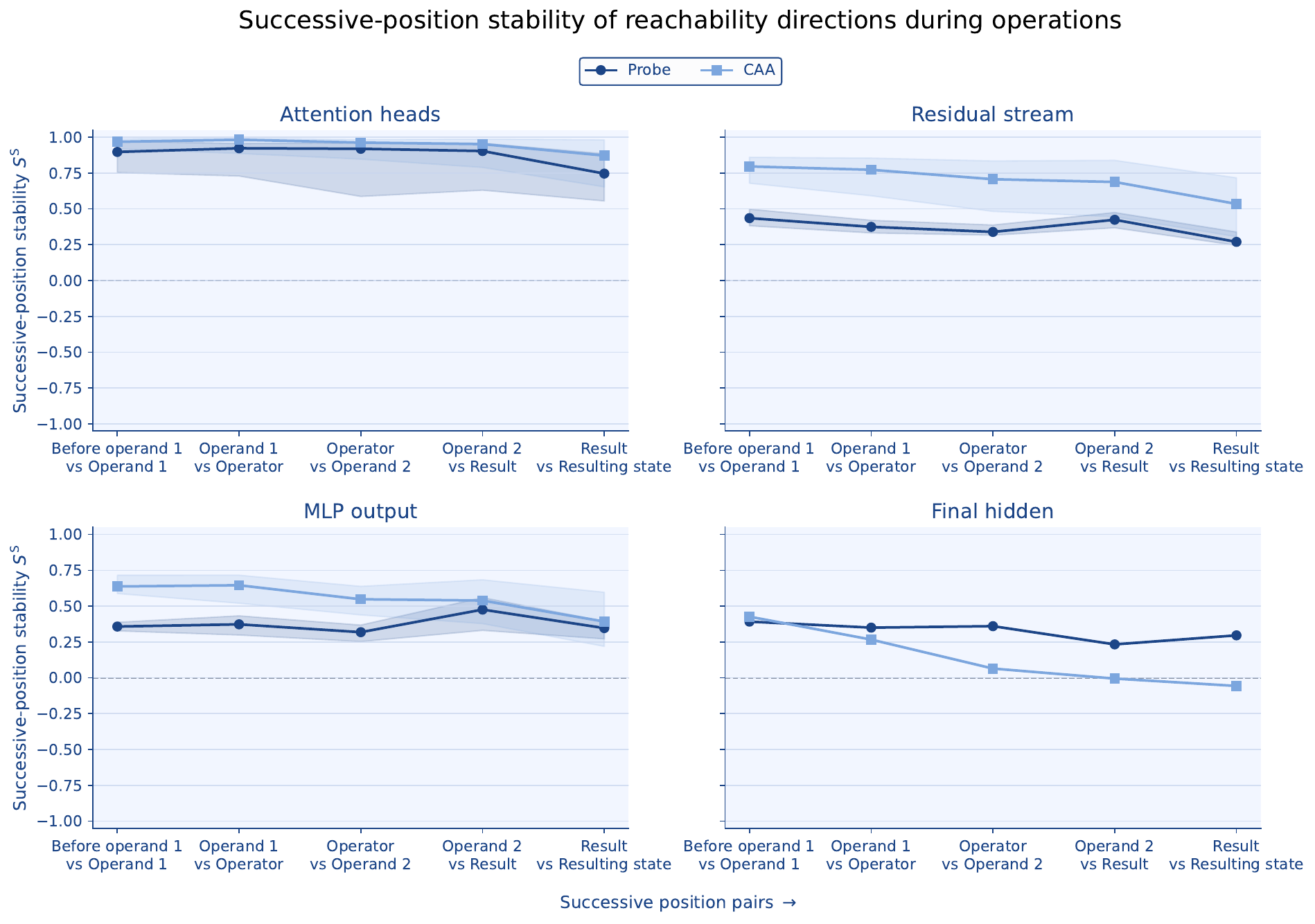}
\vspace{-1em}
\caption{We measure successive-position stability
(\(S^{\mathrm{S}}\)) to examine how \textbf{\underline{reachability}} directions
change as an arithmetic state transition is constructed. Probe
and CAA reachability directions remain similar across
successive token positions at attention heads, with lower
stability in other model components. Successive-position
stability for CAA directions at the final hidden
representation approaches zero at later transitions.
Lines show medians across locations within each component
class, and shading shows interquartile ranges.}
\label{fig:successive_position_stability_operation_reachability}
\end{figure} 

\begin{figure}[H]
\centering
\vspace{0.2em}
\includegraphics[width=0.8\linewidth]{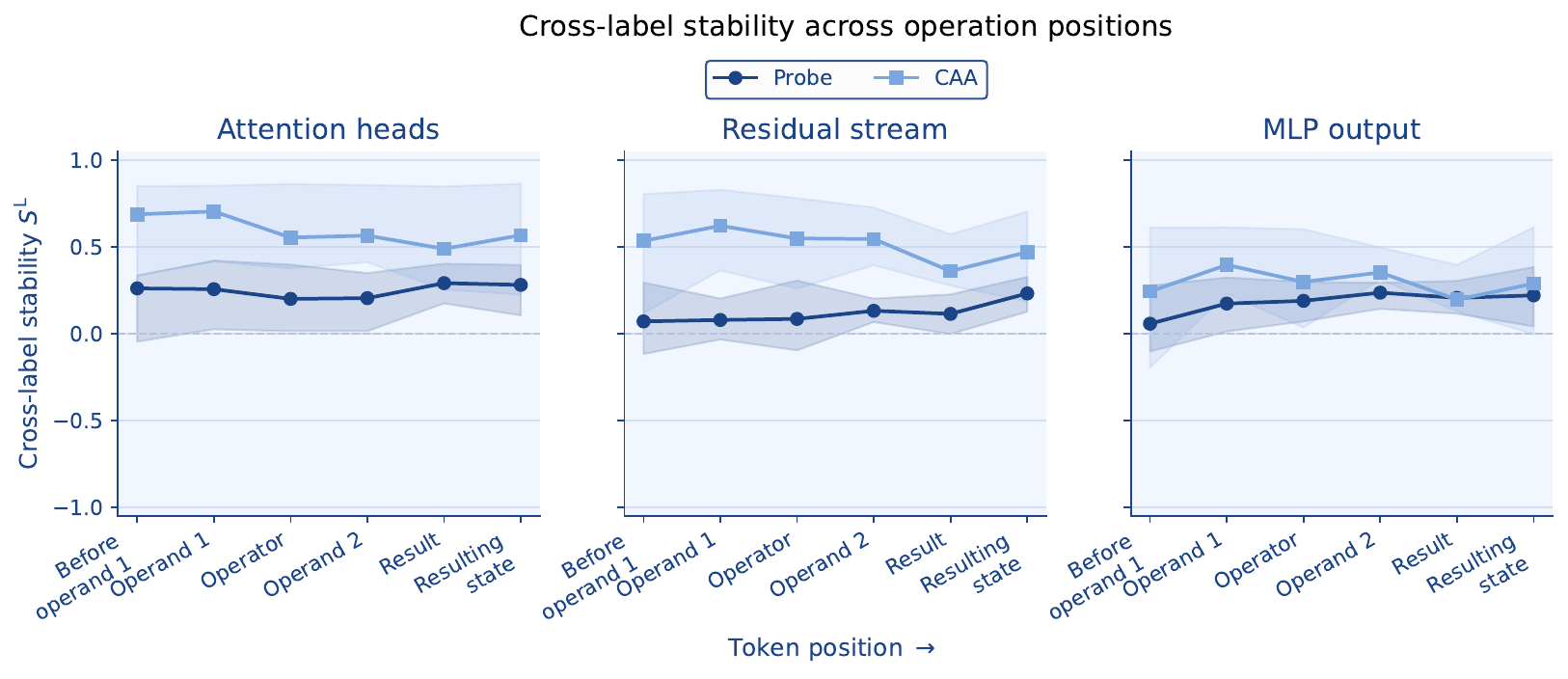}
\vspace{-1em}
\caption{Cross-label stability (\(S^{\mathrm{L}}\))
measures whether next-state reachability and final-answer
correctness are associated with similar directions at
the same token position and model location. CAA generally
yields higher cross-label stability than Probe, particularly
at attention heads and the residual stream, although the
two correctness labels do not share an identical direction.
Lines show median cosine similarity within each component
class, and shading shows interquartile ranges.}
\label{fig:cross_label_stability_operation}
\end{figure}

\clearpage
\vspace{-1em}
\section{Additional Causal Steering Analyses}
\label{sec:additional_causal_steering}
\subsection{Additional Attention-Head Interventions}
\label{sec:additional_attention_head_interventions} 

\begin{figure}[H]
\centering
\vspace{2.0em} 
\includegraphics[width=0.95\linewidth]{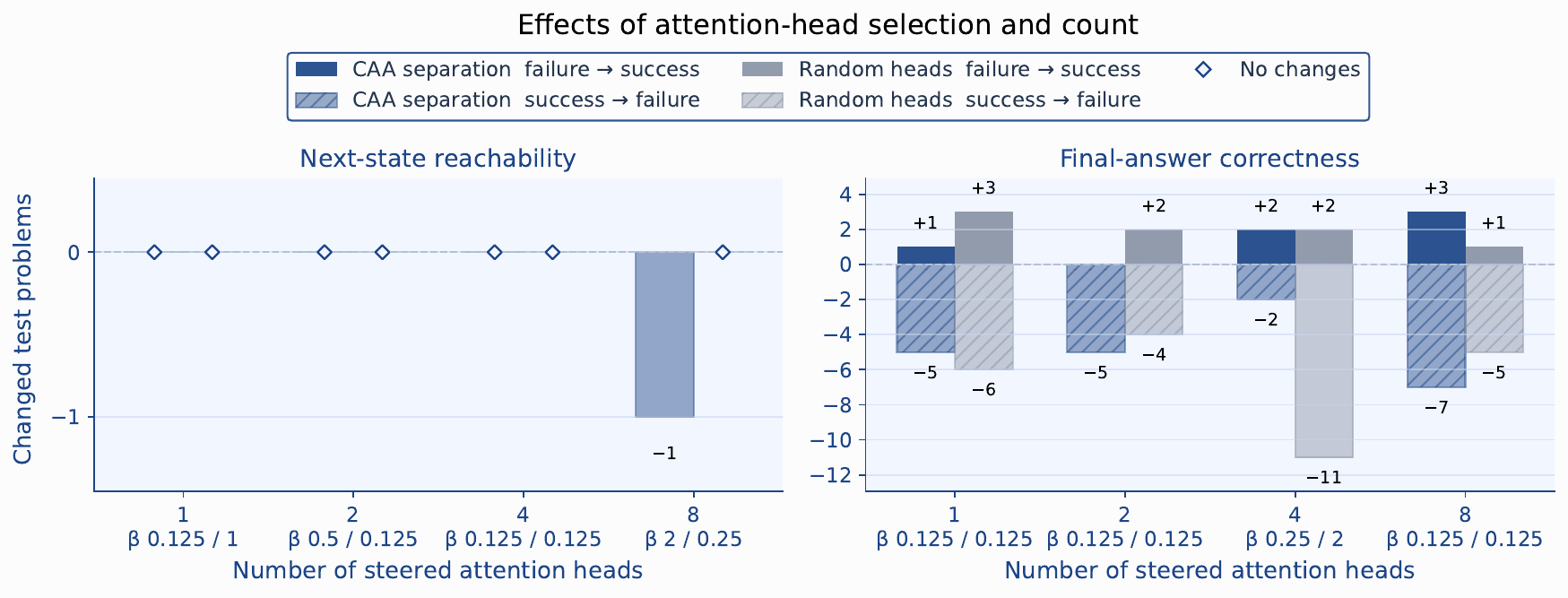}
\vspace{-0.5em}
\caption{We vary the number of steered heads and compare selection by CAA separation with random selection, measuring changes in next-state reachability and final-answer correctness. Below each head count, the two \(\beta\) values are chosen using validation data and shown as ``\texttt{\small CAA selection / random selection}''. Each setting is evaluated on 512 test problems, with and without steering. Selecting heads by CAA separation gives no consistent advantage over random selection. }
\label{fig:sparse_attention_head_heldout_counts}
\end{figure}

\begin{figure}[H]
\centering
\vspace{2.8em}
\includegraphics[width=0.95\linewidth]{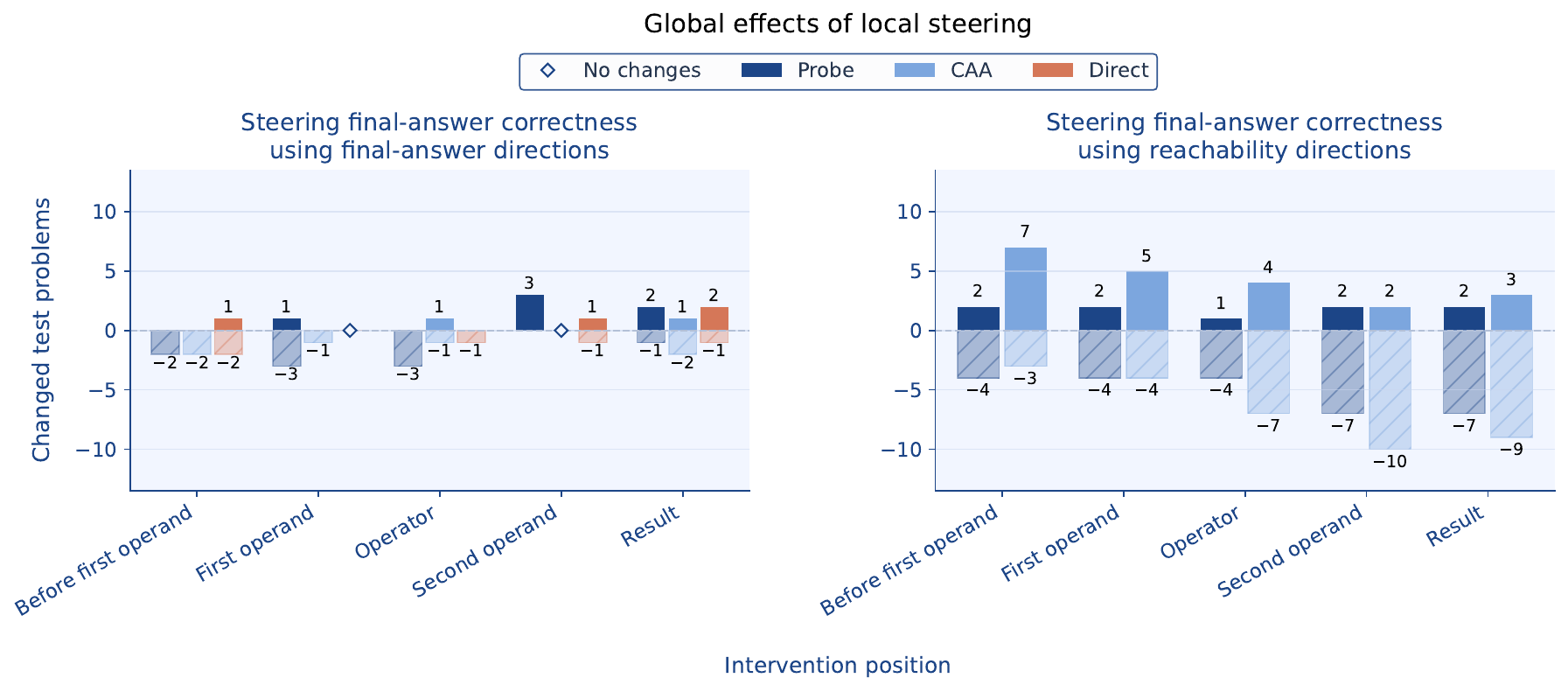}
\vspace{-1.0em}
\caption{We compare whether final-answer correctness improves when steering uses directions constructed for final-answer correctness (left) or next-state reachability (right). Neither type of direction produces consistently beneficial effects across intervention positions. All interventions use four heads, with steering strengths chosen separately using validation data. Each setting is evaluated on 512 test problems, with and without steering.}
\label{fig:heldout_position_local_global_counts}
\end{figure}

\clearpage

\begin{figure}[H]
\centering
\vspace{1.2em}
\includegraphics[width=0.95\linewidth]{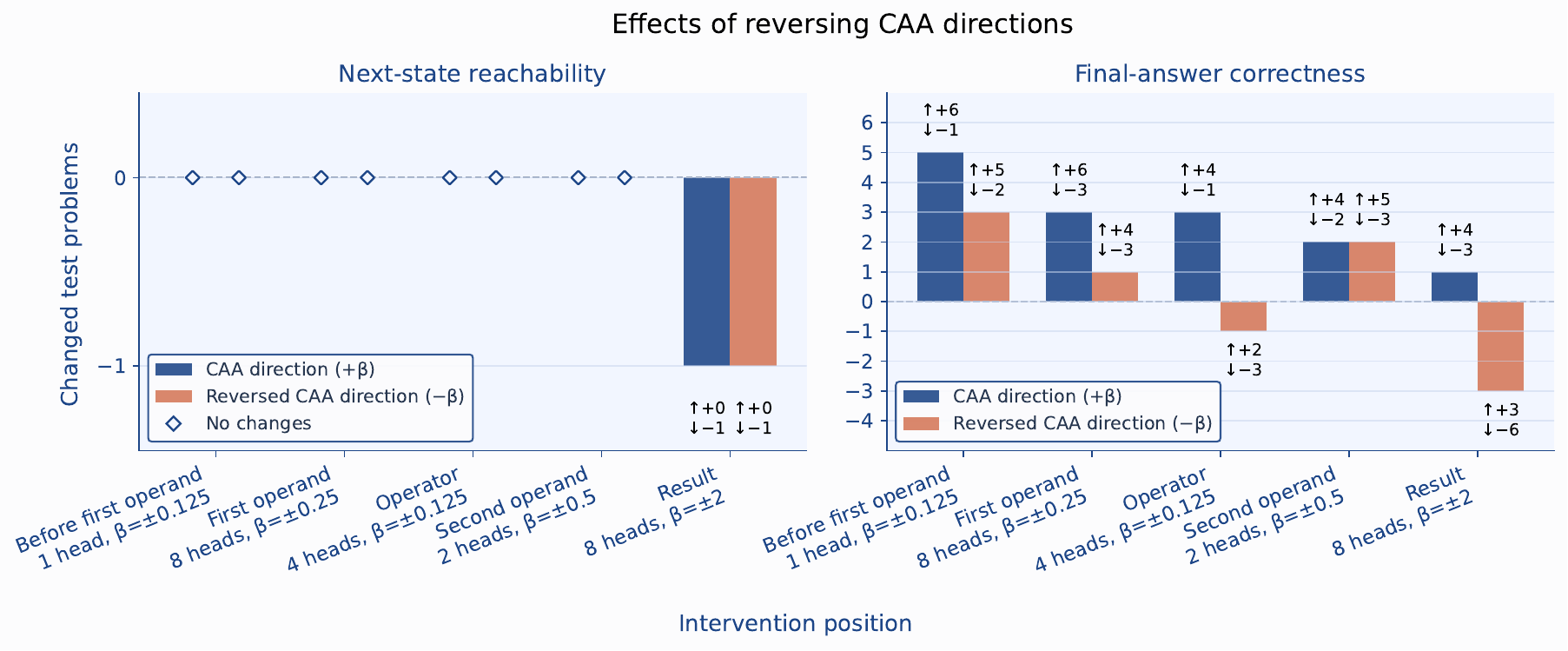}
\vspace{-1.0em}
\caption{We reverse the CAA sign chosen using validation data to test whether the opposite direction reverses its effect on next-state reachability and final-answer correctness. Bar heights show the number of problems improved minus the number worsened, with annotations reporting both counts. Neither sign consistently outperforms the other. Each sign is evaluated on the same 512 test problems and compared with generation without steering.}
\label{fig:attention_head_directionality_counts}
\end{figure}


\vspace{2em}
\subsection{Broader Steering Interventions Beyond Attention Heads}
\label{sec:interventions_beyond_attention_heads}

\clearpage
\begin{figure}[!ht]
\centering
\vspace{1.2em}
\begin{subfigure}[!h]{1.0\linewidth}
\centering
\includegraphics[width=\linewidth]
{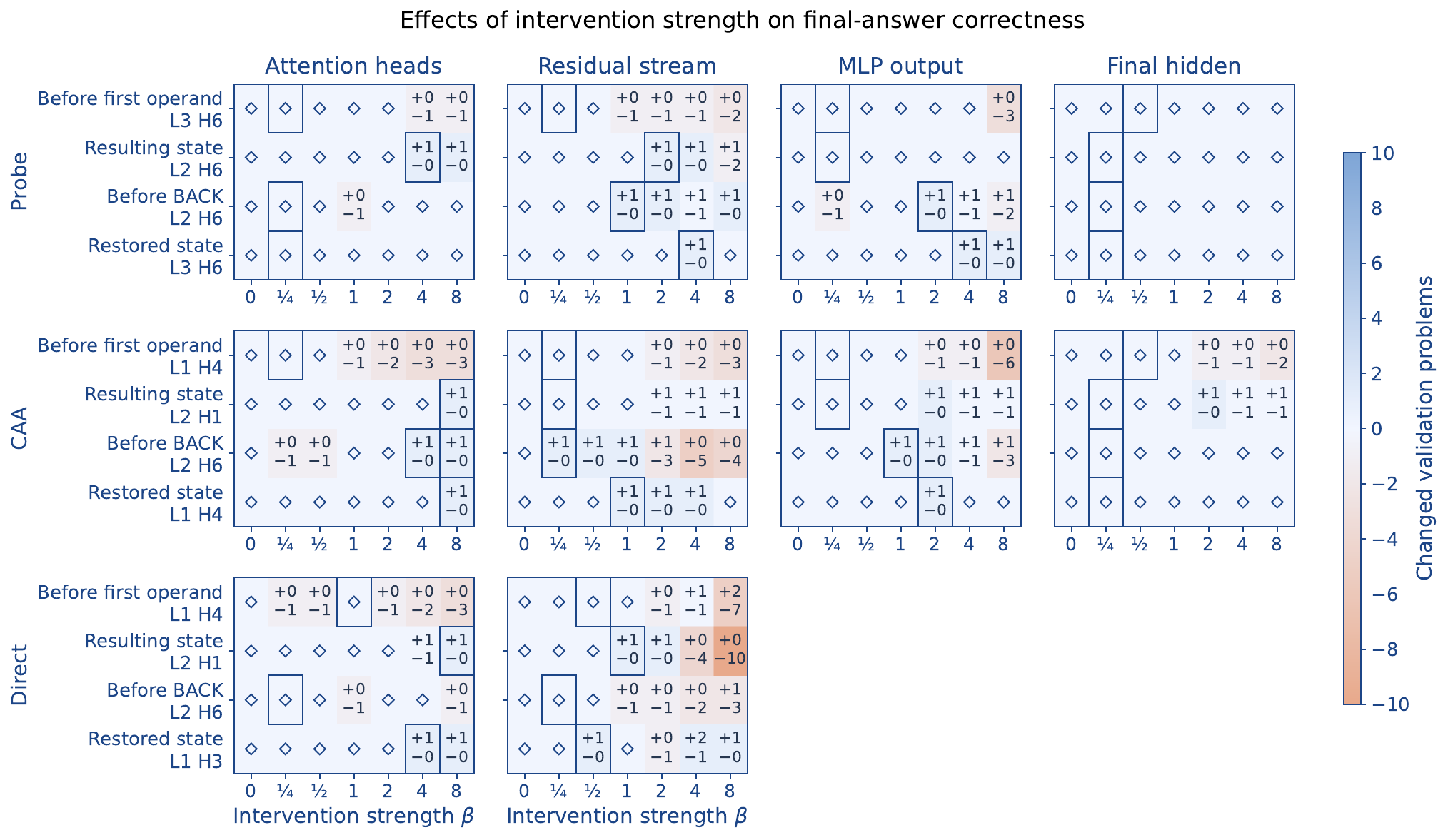}
\vspace{-0.5em}
\caption{}
\label{fig:where_and_when_dose_response_final_answer}
\end{subfigure}
\begin{subfigure}[!h]{1.0\linewidth}
\centering
\vspace{-0.0em}
\includegraphics[width=\linewidth]
{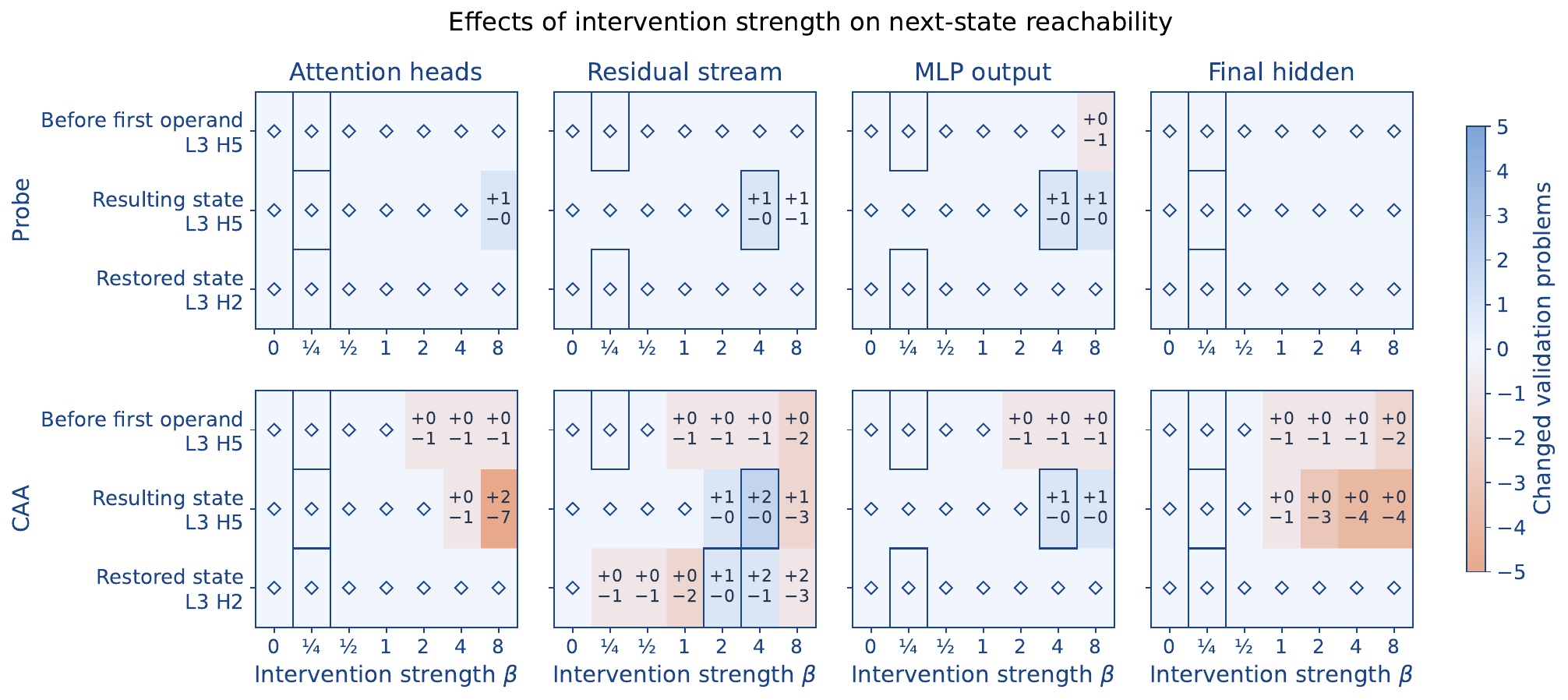}
\vspace{-0.5em}
\caption{}
\label{fig:where_and_when_dose_response_reachability}
\end{subfigure}
\vspace{-0.4em}
\caption{Varying steering strength \(\beta\) does not reveal a consistently beneficial response in either final-answer correctness (a) or next-state reachability (b). Larger interventions often change more outcomes, but some are improved and others worsened. Each cell reports the numbers of problems improved and worsened, and outlined cells mark strengths chosen using validation data. Each setting is evaluated on 16 validation problems, with and without steering.}
\label{fig:where_and_when_dose_response}
\end{figure}

\clearpage

\begin{figure}[H]
\centering
\vspace{10em}
\includegraphics[width=0.92\linewidth]{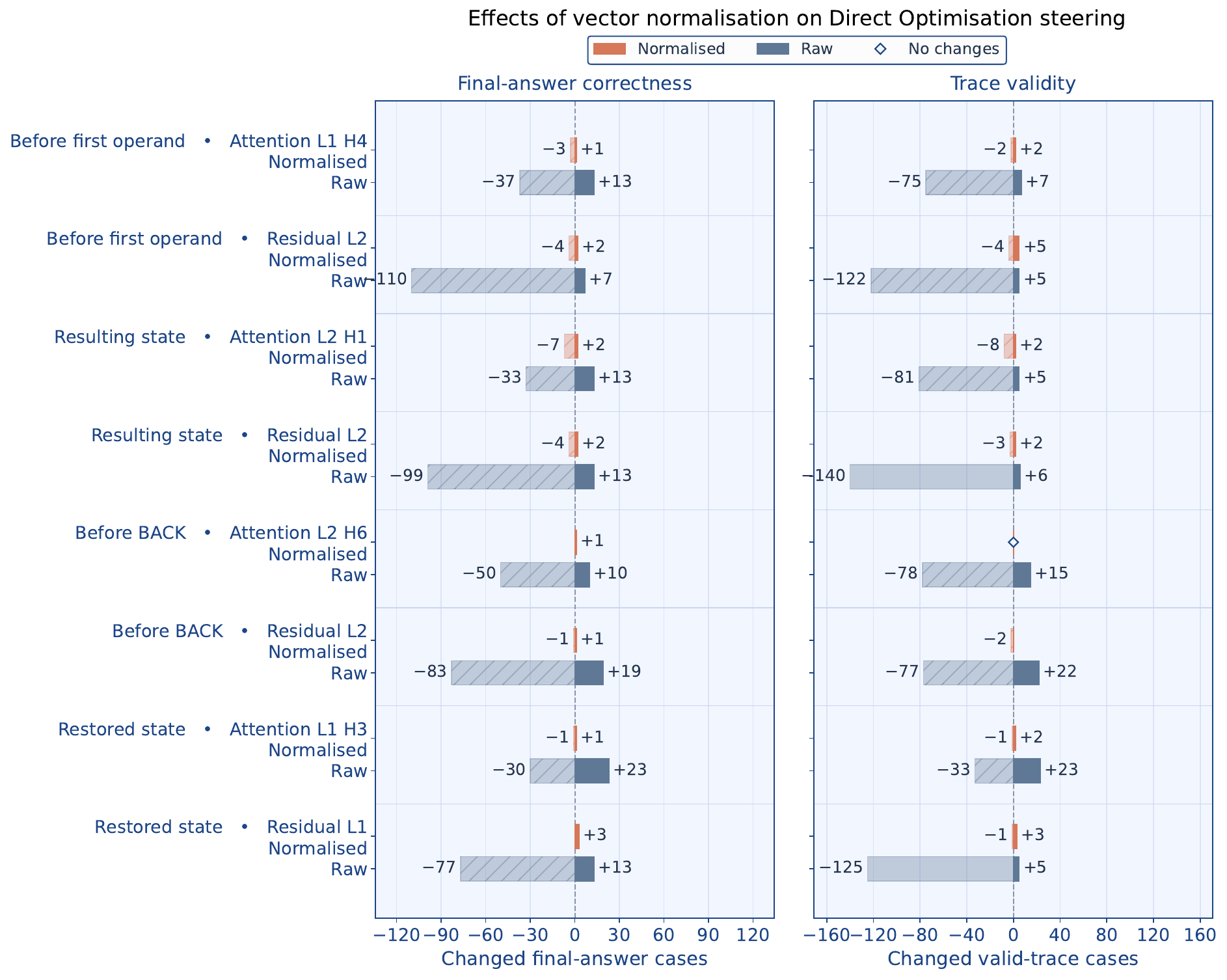}
\vspace{-0.6em}
\caption{Applying the vector learned through Direct Optimisation without normalising it changes more outcomes than applying its version normalised to unit length, but more outcomes are worsened than improved for both final-answer correctness and trace validity in these settings. We compare the two versions at the same intervention position and model component. Each version is evaluated on the same 256 test problems, with and without steering.}
\label{fig:steering_native_direct_pairs}
\end{figure}

\raggedbottom
\clearpage
\section{Full List of Heuristics}
\label{sec:heuristics_list}
{
In this section, please note that ``reward'' is to be understood as a reduction in heuristic cost.
\begin{heuristicslist}

\heuristicitem{Target-Driven Lookahead}
This family of heuristics begins with the target number, working backward to decompose the problem or prune the search space before any calculations are made.

\heuristicsubitem{The Difference Method}

\noindent\textit{Strategy Description:} This is the dominant procedural pattern, embodying a \texttt{`}Multiply Early, Adjust Late' game plan. A player performs a large multiplication to create an anchor number close to the target, then reframes the problem into solving for the small remaining difference \citep{apterous_techniques, defays1995numbo, datagenetics_countdown}.

\noindent\textit{Potential Reward Signal:} A positive reward proportional to the reduction in absolute distance between an intermediate result and the final target.

\heuristicsubitem{Quotient Foresight}

\noindent\textit{Strategy Description:} A related lookahead tactic where players mentally check if the target is roughly a multiple of an available number. They assess whether the integer quotient `(\texttt{Target / x})' is easily constructible from the remaining numbers before committing to a path \citep{defays1995numbo, apterous_techniques}.

\noindent\textit{Potential Reward Signal:} A bonus for creating a number that is an exact quotient, or for moving into a state where the new subproblem is solvable.

\heuristicsubitem{Residue \texorpdfstring{\curlyamp}{\&} Parity Pruning}

\noindent\textit{Strategy Description:} An advanced tactic where players use number properties to prune impossible solution paths. Instead of calculating a full result, they use fast mental shortcuts to check for plausibility. Common techniques include:
\begin{itemize}[noitemsep,leftmargin=*,topsep=0pt,partopsep=0pt]
\item {Parity (Odd/Even):} If the target is odd, the final arithmetic operation cannot be even * anything, even + even, or even - even, eliminating many branches.
\item {Last Digit (mod 10):} The target's final digit constrains the final operation. A target ending in \texttt{}3' guides the search towards combinations that can produce a 3.
\item {Large Number Residue (mod 25):} For a target like 783, a player might recognize 783 = 25 * 31 + 8, reframing the problem into a subgoal of making the remainder 8.
\end{itemize}
These checks allow a player to quickly discard entire subtrees of the search space without performing the full calculations \citep{reddit_six_small}.

\noindent\textit{Potential Reward Signal:} A multi-component bonus awarded for actions that align with the target's properties:
\begin{itemize}[noitemsep,leftmargin=*,topsep=0pt,partopsep=0pt]
\item {Parity Match:} A small bonus if an intermediate result has the same parity (odd/even) as the final target.
\item {Last Digit Match:} A small bonus if an arithmetic operation produces a result whose last digit matches the target's last digit.
\item {Remainder Subgoal:} A more significant bonus for creating an intermediate number that is exactly equal to the target's remainder when divided by a large number (25, 50, or 75) that is currently in the state.
\end{itemize}

\heuristicitem{Structural Decomposition}
Rather than a linear path, these strategies involve breaking the problem itself into more manageable, independent parts.

\heuristicsubitem{Decomposing the Target (Factorization)}

\noindent\textit{Strategy Description:} A powerful divide-and-conquer tactic where the target is mentally decomposed into its factors (e.g., \texttt{744 = 8 x 93}). These factors then become simpler, intermediate sub-goals to be solved \citep{apterous_techniques, defays1995numbo}.

\noindent\textit{Potential Reward Signal:} A positive reward for creating an intermediate number that is a nontrivial factor of the target number.

\heuristicsubitem{Decomposing the Numbers (Partition-and-Combine)}

\noindent\textit{Strategy Description:} A structural pattern where the player mentally splits the available numbers into two groups, solves for two intermediate subresults, and then combines them with a final operation, such as `(a+b) * (c+d)' \citep{alliot2015final, defays1995numbo}.

\noindent\textit{Potential Reward Signal:} A positive reward if two separate intermediate results in the state can be combined to directly reach the target.

\heuristicitem{Constructive Planning \texorpdfstring{\curlyamp}{\&} Waypoint Creation}
\noindent This family of heuristics covers the idea that expert players don't just use the numbers they are given; they actively construct the numbers they need.

\heuristicsubitem{Building Operands (Legal Decomposition)}

\noindent\textit{Strategy Description:} A constructive tactic where players build a required operand on the fly. This is a legal application of the distributive property, such as forming a multiplier by first summing two smaller numbers, e.g., finding \texttt{x * (a + b)}. This demonstrates multi-level planning, where one calculation serves as a subroutine for another \citep{apterous_techniques}.

\noindent\textit{Potential Reward Signal:} A positive reward for using the result of one arithmetic operation as an operand in an immediate subsequent multiplication.

\heuristicsubitem{Building Waypoints (High-Divisor Composites)}

\noindent\textit{Strategy Description:} A broader strategy where players deliberately form cognitively easy and mathematically flexible waypoints. Expert players learn to prefer creating ``high-divisor composites'' (e.g., 60, 72, 90) over large primes, as these numbers act as powerful hubs with many factors for subsequent steps \citep{reddit_six_small, defays1995numbo, datagenetics_countdown}.

\noindent\textit{Potential Reward Signal:} A small bonus for hitting waypoints from a predefined list of "useful composites," especially if they reduce the distance to the target.

\heuristicsubitem{Strategic Doubling and Halving}

\noindent\textit{Strategy Description:} Players treat certain numbers as functional tools for rapid scaling. Numbers like 2 and 4 are used for power-of-two scaling to quickly approach the target's order of magnitude. A common application is the ``half-difference trick,'' where a player creates a state where the remaining gap to the target can be easily bridged by a final doubling or halving calculation \citep{apterous_techniques}.

\noindent\textit{Potential Reward Signal:} A small positive reward for any multiplication or division by 2 or 4 that reduces the distance to the target.

\heuristicitem{Strategic Foresight \texorpdfstring{\curlyamp}{\&} Resource Management}
These advanced strategies require looking ahead to evaluate moves based on their future consequences.

\heuristicsubitem{Preserving High-Value Pairs (Partner Preservation)}

\noindent\textit{Strategy Description:} A sophisticated strategy of preserving "canonical partners" for high-impact moves (e.g., saving a \texttt{4} to be used with a 25). This requires evaluating moves based on their opportunity cost \citep{apterous_techniques, reddit_six_small}.

\noindent\textit{Potential Reward Signal:} A small negative reward (cost) for using a canonical partner in a low-value arithmetic operation (e.g., addition).

\heuristicsubitem{Maintaining State Flexibility (Composite Preference)}

\noindent\textit{Strategy Description:} A key aspect of strategic foresight is maintaining a flexible board state by preferring to create intermediate numbers that are themselves easy to decompose. Expert players develop an intuitive preference for "high-divisor composites" (numbers with many small, obvious factors, e.g., 60, 72, 90) as these act as powerful and flexible hubs for subsequent arithmetic operations. Conversely, this leads to an implicit strategy of "Early Prime-Avoidance" \citep{reddit_six_small}.

\noindent\textit{Potential Reward Signal:} A small negative reward (cost) for generating a large intermediate number that is not divisible by 2, 3, or 5, unless it is already very close to the target.

\heuristicitem{Meta-Gaming \texorpdfstring{\curlyamp}{\&} Efficiency}

\noindent Finally, expert play is shaped by the rules of the game and the cognitive limits of the player.

\heuristicsubitem{Adapting to Scoring Rules (Near-Miss Strategy)}

\noindent\textit{Strategy Description:} A pragmatic shift from an optimization objective (find the exact target) to a satisficing one (land within the \texttt{±10} scoring range) when an exact solution seems intractable under time pressure \citep{wikipedia_countdown_game_show, apterous_techniques}.

\noindent\textit{Potential Reward Signal:} In a multi-objective setting, a secondary reward for terminating within the \texttt{±10} scoring range.

\heuristicsubitem{Optimizing for Simplicity (Step \texorpdfstring{\curlyamp}{\&} Identity Efficiency)}

\noindent\textit{Strategy Description:} Skilled players prefer shorter solutions and avoid ``wasted'' moves that offer no progress, such as adding or subtracting zero, or multiplying or dividing by one. This cognitive preference for efficiency is also a key metric for optimality in algorithmic solvers \citep{alliot2015final, datagenetics_countdown}.

\noindent\textit{Potential Reward Signal:} A small, per-step penalty to encourage efficiency, and a larger penalty for generating a useless identity calculation.

\end{heuristicslist}
}


\begin{SaveVerbatim}[commandchars=\\\{\}]{chaintrace_back_one}
\underline{Problem}
IN: [ 1, 1, 2, 9, 25, 50 ] TGT: 675

\underline{Solution}
[ 1, 1, 2, 9, 25, 50 ] -> 9 * 50 = 450
-> [ 1, 1, 2, 25, 450 ] ;
BACK -> [ 1, 1, 2, 9, 25, 50 ] ;
[ 1, 1, 2, 9, 25, 50 ] -> 25 + 50 = 75
-> [ 1, 1, 2, 9, 75 ] ;
[ 1, 1, 2, 9, 75 ] -> 9 * 75 = 675
-> [ 1, 1, 2, 675 ] ;
SUCCESS ;
EXPR ((9 * (25 + 50))) \raisedcheckmark
\end{SaveVerbatim}

\begin{SaveVerbatim}[commandchars=\\\{\}]{chaintrace_back_two}
\underline{Problem}
IN: [ 3, 3, 9, 9, 25, 50 ] TGT: 275

\underline{Solution}
[ 3, 3, 9, 9, 25, 50 ] -> 3 * 50 = 150
-> [ 3, 9, 9, 25, 150 ] ;
BACK -> [ 3, 3, 9, 9, 25, 50 ] ;
[ 3, 3, 9, 9, 25, 50 ] -> 9 * 50 = 450
-> [ 3, 3, 9, 25, 450 ] ;
BACK -> [ 3, 3, 9, 9, 25, 50 ] ;
[ 3, 3, 9, 9, 25, 50 ] -> 9 * 25 = 225
-> [ 3, 3, 9, 50, 225 ] ;
[ 3, 3, 9, 50, 225 ] -> 50 + 225 = 275
-> [ 3, 3, 9, 275 ] ;
SUCCESS ;
EXPR ((50 + (9 * 25))) \raisedcheckmark
\end{SaveVerbatim}

\begin{SaveVerbatim}[commandchars=\\\{\}]{chaintrace_back_three}
\underline{Problem}
IN: [ 5, 7, 7, 9, 25, 100 ] TGT: 108

\underline{Solution}
[ 5, 7, 7, 9, 25, 100 ] -> 7 + 100 = 107
-> [ 5, 7, 9, 25, 107 ] ;
BACK -> [ 5, 7, 7, 9, 25, 100 ] ;
[ 5, 7, 7, 9, 25, 100 ] -> 5 + 100 = 105
-> [ 7, 7, 9, 25, 105 ] ;
BACK -> [ 5, 7, 7, 9, 25, 100 ] ;
[ 5, 7, 7, 9, 25, 100 ] -> 9 + 100 = 109
-> [ 5, 7, 7, 25, 109 ] ;
BACK -> [ 5, 7, 7, 9, 25, 100 ] ;
[ 5, 7, 7, 9, 25, 100 ] -> 5 + 7 = 12
-> [ 7, 9, 12, 25, 100 ] ;
[ 7, 9, 12, 25, 100 ] -> 9 * 12 = 108
-> [ 7, 25, 100, 108 ] ;
SUCCESS ;
EXPR ((9 * (5 + 7))) \raisedcheckmark
\end{SaveVerbatim}

\begin{SaveVerbatim}[commandchars=\\\{\}]{chaintrace_back_four}
\underline{Problem}
IN: [ 1, 5, 9, 9, 25, 75 ] TGT: 120

\underline{Solution}
[ 1, 5, 9, 9, 25, 75 ] -> 25 + 75 = 100
-> [ 1, 5, 9, 9, 100 ] ;
BACK -> [ 1, 5, 9, 9, 25, 75 ] ;
[ 1, 5, 9, 9, 25, 75 ] -> 9 + 75 = 84
-> [ 1, 5, 9, 25, 84 ] ;
BACK -> [ 1, 5, 9, 9, 25, 75 ] ;
[ 1, 5, 9, 9, 25, 75 ] -> 5 * 25 = 125
-> [ 1, 9, 9, 75, 125 ] ;
BACK -> [ 1, 5, 9, 9, 25, 75 ] ;
[ 1, 5, 9, 9, 25, 75 ] -> 5 + 75 = 80
-> [ 1, 9, 9, 25, 80 ] ;
BACK -> [ 1, 5, 9, 9, 25, 75 ] ;
[ 1, 5, 9, 9, 25, 75 ] -> 25 - 1 = 24
-> [ 5, 9, 9, 24, 75 ] ;
[ 5, 9, 9, 24, 75 ] -> 5 * 24 = 120
-> [ 9, 9, 75, 120 ] ;
SUCCESS ;
EXPR ((5 * (25 - 1))) \raisedcheckmark
\end{SaveVerbatim}

\begin{SaveVerbatim}[commandchars=\\\{\}]{chaintrace_back_five}
\underline{Problem}
IN: [ 2, 3, 3, 5, 9, 50 ] TGT: 135

\underline{Solution}
[ 2, 3, 3, 5, 9, 50 ] -> 3 * 50 = 150
-> [ 2, 3, 5, 9, 150 ] ;
BACK -> [ 2, 3, 3, 5, 9, 50 ] ;
[ 2, 3, 3, 5, 9, 50 ] -> 2 * 50 = 100
-> [ 3, 3, 5, 9, 100 ] ;
BACK -> [ 2, 3, 3, 5, 9, 50 ] ;
[ 2, 3, 3, 5, 9, 50 ] -> 3 * 50 = 150
-> [ 2, 3, 5, 9, 150 ] ;
[ 2, 3, 5, 9, 150 ] -> 150 - 9 = 141
-> [ 2, 3, 5, 141 ] ;
BACK -> [ 2, 3, 5, 9, 150 ] ;
[ 2, 3, 5, 9, 150 ] -> 150 - 5 = 145
-> [ 2, 3, 9, 145 ] ;
BACK -> [ 2, 3, 5, 9, 150 ] ;
[ 2, 3, 5, 9, 150 ] -> 150 - 3 = 147
-> [ 2, 5, 9, 147 ] ;
BACK -> [ 2, 3, 3, 5, 9, 50 ] ;
[ 2, 3, 3, 5, 9, 50 ] -> 5 * 9 = 45
-> [ 2, 3, 3, 45, 50 ] ;
[ 2, 3, 3, 45, 50 ] -> 3 * 45 = 135
-> [ 2, 3, 50, 135 ] ;
SUCCESS ;
EXPR ((3 * (5 * 9))) \raisedcheckmark
\end{SaveVerbatim}

\begin{SaveVerbatim}[commandchars=\\\{\}]{chaintrace_back_six}
\underline{Problem}
IN: [ 3, 7, 7, 8, 10, 75 ] TGT: 248

\underline{Solution}
[ 3, 7, 7, 8, 10, 75 ] -> 3 * 75 = 225
-> [ 7, 7, 8, 10, 225 ] ;
BACK -> [ 3, 7, 7, 8, 10, 75 ] ;
[ 3, 7, 7, 8, 10, 75 ] -> 3 + 75 = 78
-> [ 7, 7, 8, 10, 78 ] ;
BACK -> [ 3, 7, 7, 8, 10, 75 ] ;
[ 3, 7, 7, 8, 10, 75 ] -> 3 * 75 = 225
-> [ 7, 7, 8, 10, 225 ] ;
[ 7, 7, 8, 10, 225 ] -> 10 + 225 = 235
-> [ 7, 7, 8, 235 ] ;
BACK -> [ 7, 7, 8, 10, 225 ] ;
[ 7, 7, 8, 10, 225 ] -> 7 + 225 = 232
-> [ 7, 8, 10, 232 ] ;
BACK -> [ 3, 7, 7, 8, 10, 75 ] ;
[ 3, 7, 7, 8, 10, 75 ] -> 8 * 10 = 80
-> [ 3, 7, 7, 75, 80 ] ;
[ 3, 7, 7, 75, 80 ] -> 3 * 80 = 240
-> [ 7, 7, 75, 240 ] ;
BACK -> [ 3, 7, 7, 8, 10, 75 ] ;
[ 3, 7, 7, 8, 10, 75 ] -> 3 * 75 = 225
-> [ 7, 7, 8, 10, 225 ] ;
[ 7, 7, 8, 10, 225 ] -> 8 + 225 = 233
-> [ 7, 7, 10, 233 ] ;
BACK -> [ 3, 7, 7, 8, 10, 75 ] ;
[ 3, 7, 7, 8, 10, 75 ] -> 10 + 75 = 85
-> [ 3, 7, 7, 8, 85 ] ;
[ 3, 7, 7, 8, 85 ] -> 3 * 85 = 255
-> [ 7, 7, 8, 255 ] ;
[ 7, 7, 8, 255 ] -> 255 - 7 = 248
-> [ 7, 8, 248 ] ;
SUCCESS ;
EXPR (((3 * (10 + 75)) - 7)) \raisedcheckmark
\end{SaveVerbatim}

\begin{SaveVerbatim}[commandchars=\\\{\}]{chaintrace_back_seven}
\underline{Problem}
IN: [ 1, 2, 5, 7, 8, 75 ] TGT: 515

\underline{Solution}
[ 1, 2, 5, 7, 8, 75 ] -> 7 * 75 = 525
-> [ 1, 2, 5, 8, 525 ] ;
BACK -> [ 1, 2, 5, 7, 8, 75 ] ;
[ 1, 2, 5, 7, 8, 75 ] -> 8 * 75 = 600
-> [ 1, 2, 5, 7, 600 ] ;
BACK -> [ 1, 2, 5, 7, 8, 75 ] ;
[ 1, 2, 5, 7, 8, 75 ] -> 2 * 75 = 150
-> [ 1, 5, 7, 8, 150 ] ;
BACK -> [ 1, 2, 5, 7, 8, 75 ] ;
[ 1, 2, 5, 7, 8, 75 ] -> 5 * 75 = 375
-> [ 1, 2, 7, 8, 375 ] ;
BACK -> [ 1, 2, 5, 7, 8, 75 ] ;
[ 1, 2, 5, 7, 8, 75 ] -> 7 * 75 = 525
-> [ 1, 2, 5, 8, 525 ] ;
[ 1, 2, 5, 8, 525 ] -> 525 - 8 = 517
-> [ 1, 2, 5, 517 ] ;
BACK -> [ 1, 2, 5, 7, 8, 75 ] ;
[ 1, 2, 5, 7, 8, 75 ] -> 5 + 75 = 80
-> [ 1, 2, 7, 8, 80 ] ;
BACK -> [ 1, 2, 5, 7, 8, 75 ] ;
[ 1, 2, 5, 7, 8, 75 ] -> 7 * 75 = 525
-> [ 1, 2, 5, 8, 525 ] ;
[ 1, 2, 5, 8, 525 ] -> 525 - 2 = 523
-> [ 1, 5, 8, 523 ] ;
BACK -> [ 1, 2, 5, 8, 525 ] ;
[ 1, 2, 5, 8, 525 ] -> 525 - 8 = 517
-> [ 1, 2, 5, 517 ] ;
[ 1, 2, 5, 517 ] -> 517 - 2 = 515
-> [ 1, 5, 515 ] ;
SUCCESS ;
EXPR ((((7 * 75) - 8) - 2)) \raisedcheckmark
\end{SaveVerbatim}

\begin{SaveVerbatim}[commandchars=\\\{\}]{chaintrace_back_eight}
\underline{Problem}
IN: [ 4, 6, 7, 9, 75, 100 ] TGT: 193

\underline{Solution}
[ 4, 6, 7, 9, 75, 100 ] -> 75 + 100 = 175
-> [ 4, 6, 7, 9, 175 ] ;
[ 4, 6, 7, 9, 175 ] -> 9 + 175 = 184
-> [ 4, 6, 7, 184 ] ;
BACK -> [ 4, 6, 7, 9, 175 ] ;
[ 4, 6, 7, 9, 175 ] -> 7 + 175 = 182
-> [ 4, 6, 9, 182 ] ;
BACK -> [ 4, 6, 7, 9, 175 ] ;
[ 4, 6, 7, 9, 175 ] -> 6 + 175 = 181
-> [ 4, 7, 9, 181 ] ;
BACK -> [ 4, 6, 7, 9, 175 ] ;
[ 4, 6, 7, 9, 175 ] -> 4 + 175 = 179
-> [ 6, 7, 9, 179 ] ;
BACK -> [ 4, 6, 7, 9, 75, 100 ] ;
[ 4, 6, 7, 9, 75, 100 ] -> 9 + 100 = 109
-> [ 4, 6, 7, 75, 109 ] ;
BACK -> [ 4, 6, 7, 9, 75, 100 ] ;
[ 4, 6, 7, 9, 75, 100 ] -> 7 + 100 = 107
-> [ 4, 6, 9, 75, 107 ] ;
BACK -> [ 4, 6, 7, 9, 75, 100 ] ;
[ 4, 6, 7, 9, 75, 100 ] -> 6 + 100 = 106
-> [ 4, 7, 9, 75, 106 ] ;
[ 4, 7, 9, 75, 106 ] -> 4 + 106 = 110
-> [ 7, 9, 75, 110 ] ;
BACK -> [ 4, 6, 7, 9, 75, 100 ] ;
[ 4, 6, 7, 9, 75, 100 ] -> 4 + 100 = 104
-> [ 6, 7, 9, 75, 104 ] ;
BACK -> [ 4, 6, 7, 9, 75, 100 ] ;
[ 4, 6, 7, 9, 75, 100 ] -> 4 * 75 = 300
-> [ 6, 7, 9, 100, 300 ] ;
[ 6, 7, 9, 100, 300 ] -> 300 - 100 = 200
-> [ 6, 7, 9, 200 ] ;
[ 6, 7, 9, 200 ] -> 200 - 7 = 193
-> [ 6, 9, 193 ] ;
SUCCESS ;
EXPR ((((4 * 75) - 100) - 7)) \raisedcheckmark
\end{SaveVerbatim}

\clearpage
\section{Additional ChainSearch Trace Examples}
\label{sec:additional_trace_examples}
{
The examples below provide one SFT sample reasoning trace per number of \texttt{BACK} transitions, from one through eight. The traces are reference trajectories rather than unique solutions, as a problem may admit multiple valid traces.}

\vspace{1em}

\begin{chainexamplebreakable}[label={example:astar_back_one}]{An A\textsuperscript{*} reference trace with \textbf{\underline{one}} \texttt{BACK} transitions}
\UseVerbatim[fontsize=\chaintracefontsize]{chaintrace_back_one}
\end{chainexamplebreakable}

\vspace{1em}

\begin{chainexamplebreakable}[label={example:astar_back_two}]{An A\textsuperscript{*} reference trace with \textbf{\underline{two}} \texttt{BACK} transitions}
\UseVerbatim[fontsize=\chaintracefontsize]{chaintrace_back_two}
\end{chainexamplebreakable}

\vspace{1em}

\begin{chainexamplebreakable}[label={example:astar_back_three}]{An A\textsuperscript{*} reference trace with \textbf{\underline{three}} \texttt{BACK} transitions}
\UseVerbatim[fontsize=\chaintracefontsize]{chaintrace_back_three}
\end{chainexamplebreakable}

\vspace{1em}

\begin{chainexamplebreakable}[label={example:astar_back_four}]{An A\textsuperscript{*} reference trace with \textbf{\underline{four}} \texttt{BACK} transitions}
\UseVerbatim[fontsize=\chaintracefontsize]{chaintrace_back_four}
\end{chainexamplebreakable}

\vspace{2em}

\begin{chainexamplebreakable}[label={example:astar_back_five}]{An A\textsuperscript{*} reference trace with \textbf{\underline{five}} \texttt{BACK} transitions}
\UseVerbatim[fontsize=\chaintracefontsize]{chaintrace_back_five}
\end{chainexamplebreakable}

\vspace{2em}

\begin{chainexamplebreakable}[label={example:astar_back_six}]{An A\textsuperscript{*} reference trace with \textbf{\underline{six}} \texttt{BACK} transitions}
\UseVerbatim[fontsize=\chaintracefontsize]{chaintrace_back_six}
\end{chainexamplebreakable}

\vspace{2em}

\begin{chainexamplebreakable}[label={example:astar_back_seven}]{An A\textsuperscript{*} reference trace with \textbf{\underline{seven}} \texttt{BACK} transitions}
\UseVerbatim[fontsize=\chaintracefontsize]{chaintrace_back_seven}
\end{chainexamplebreakable}

\vspace{2em}

\begin{chainexamplebreakable}[label={example:astar_back_eight}]{An A\textsuperscript{*} reference trace with \textbf{\underline{eight}} \texttt{BACK} transitions}
\UseVerbatim[fontsize=\chaintracefontsize]{chaintrace_back_eight}
\end{chainexamplebreakable}

\vspace{4em}

\end{document}